\documentclass[letterpaper, 10 pt, journal, twoside]{IEEEtran}
\usepackage{times}

\let\labelindent\relax
\usepackage{enumitem}

\usepackage{multicol}
\usepackage[bookmarks=true]{hyperref}
\hypersetup{
pdftitle={Pick N Place Planning},
pdfsubject={Robotics},
pdfauthor={Devendran Shanthi and Hermans},
pdfkeywords={grasping, object placement, motion planning}
}

\IEEEoverridecommandlockouts
\usepackage{array}
\usepackage{subcaption} 
\usepackage[skip=0pt,font=small,labelfont=bf]{caption} 
\usepackage{textcomp}
\usepackage{color}
\usepackage{mathtools}
\usepackage{amsmath}
\usepackage{fixme}
\usepackage{epsfig,graphicx,amsmath,amsfonts}
\usepackage{xcolor,import}
\usepackage{transparent}
\usepackage{upgreek}
\usepackage{algorithm}
\usepackage{algpseudocode}
\DeclareMathOperator*{\argmin}{arg\,min}

\usepackage[normalem]{ulem}
\usepackage[noadjust]{cite}
\usepackage[export]{adjustbox}

\usepackage{multirow}
\graphicspath{./Figures/}
\definecolor{international_orange}{RGB}{240, 74, 0}
\newcommand{\mohan}[1]{{\noindent \textcolor[rgb]{0.0,0.0,1.0}{\small {TODO [Mohan]---}{#1}}}}
\newcommand{\tucker}[1]{{\noindent \textcolor{international_orange}{\small {TODO [Tucker]---}{#1}}}}

\usepackage{amsmath, amssymb}
\usepackage{amsfonts}
\usepackage{bm}
\usepackage[normalem]{ulem}
\newcommand{\xv}{\bm{x}} 
\newcommand{\thetav}{\bm{\theta}} 
\newcommand{\qv}{\bm{q}} 
\newcommand{\sdf}{\texttt{SDF}} 
\newcommand{\dsdf}{\texttt{DSDF}} 

\algnewcommand{\algorithmicgoto}{\textbf{go to}}%
\algnewcommand{\Goto}[1]{\algorithmicgoto~\ref{#1}}%
\algnewcommand{\Continue}{\textbf{continue}}
\algnewcommand{\Break}{\textbf{break}}

\title{Pick and Place Planning is Better than Pick Planning then Place Planning}
\author{Mohanraj Devendran Shanthi$^{1}$ and Tucker Hermans$^{1,2}$%
\thanks{Manuscript received: October, 18, 2023; Revised December, 23, 2023; Accepted January, 19, 2024.}
\thanks{This paper was recommended for publication by Editor J. Borràs Sol upon evaluation of the Associate Editor  and Reviewers' comments. This work is supported by DARPA under grant N66001-19-2-4035 and by NSF Award \#1846341.}
\thanks{$^1$ Mohanraj Devendran Shanthi and Tucker Hermans are with the Kahlert School of Computing and the Robotics Center, University of Utah, Salt Lake City, UT 84112, USA. {\tt\small\{mohanraj.devendranshanthi, tucker.hermans\}@utah.edu}}%
\thanks{$^2$ Tucker Hermans is also with NVIDIA, Seattle, WA 95050, USA.}%
\thanks{Digital Object Identifier (DOI): see top of this page.}
}

\begin{document}
\frenchspacing
\input{intro-fig}
\maketitle

\begin{abstract}
  Robotic pick and place stands at the heart of autonomous manipulation.
  When conducted in cluttered or complex environments robots must jointly reason about the selected grasp and desired placement locations to ensure success. While several works have examined this joint pick-and-place problem, none have fully leveraged recent learning-based approaches for multi-fingered grasp planning. We present a modular algorithm for joint pick and place planning that can make use of state of the art grasp classifiers for planning multi-fingered grasps for novel objects from partial view point clouds. We demonstrate our joint pick and place formulation with several costs associated with different placement tasks. Experiments on pick and place tasks with cluttered scenes using a physical robot show that our joint inference method is more successful than a sequential pick then place approach, while also achieving better placement configurations.
\end{abstract}

\vspace{40pt}
\section{Introduction}
\IEEEPARstart{P}{ick} and place operations, where a robot grasps, lifts, and then safely deposits an object at a desired location, define the quintessential problem in robotic manipulation.
The research literature reflects this key importance with considerable work examining grasping objects~\cite{grupen-icra1991-planning-grasp-strategies,Grasp-SampleBased-RAM23}, with contemporary methods capable of grasping novel objects with high success~\cite{Grasp-Hermans-Lu-VoxelInf-RAM-2020, Grasp-Leitner-2018-Closeloop-Novel-Grasping, Grasp-Mahler-DexNet2.0-2017}.
Research focused on object placement, though not as extensive as grasping, investigates various aspects including stability of placements~\cite{Place-StoneStack2017, Place-Henrich-GeometricGpuBased2014}, semantic placement~\cite{Place-SemanticPlacement-Paxton-Hermans-Fox-2021, Place-SaxenaPlaceLearn2012}, and multi-object rearrangement~\cite{Place-TableTopRearrange,cosgun-iros2011}. Though pick and place naturally go hand-in-hand, most research investigates the two highly related sub-tasks individually.

Treating the problems independently ignores a number of important issues. In particular, while grasp success is necessary for successful placement, it is not sufficient to guarantee it. In fact, a grasp configuration might succeed in lifting an object, but could end up contributing to placement failure if the robot collides with other objects in the scene during placement as shown in Fig.~(1)(Left \& Middle).
Likewise, if one ensures that a previously planned grasp does not collide with objects during placement, it might do so at the expense of object instability or reachability by the robot at the placement location Fig.~(1)(Right). Thus causing either placement planning or execution failure.

Works that have tackled pick and place jointly restrict themselves in some way, making simplifying assumptions not needed by modern grasp planners.
These simplifications include requiring full geometry of the object and environment as meshes~\cite{PnP-KragicMonteCarloPlace2019}, a restricted class of known object categories~\cite{PnP-AttentionFocusReinfLearning-Gualtieri-Platt-2018}, restricting the planner to use a fixed subset of grasps (e.g. overhead) ~\cite{PnP-SelfSupervisedGraspPlaceEmbedding-Kroger-2020}, or simplified grippers~\cite{PnP-Pick2Place2023}.

In contrast, we examine the problem of joint pick and place planning given only partial view point clouds of the object and environment. This includes the case of grasping and placing previously unseen objects. Further, we plan over arbitrary grasps from the full continuous space of feasible robot configurations, as done in recent grasping work~\cite{Grasp-Hermans-Lu-VoxelInf-RAM-2020}.

We formalize the joint pick and place task as a constrained optimization problem (Sec.~\ref{sec:probdef}). Our framework enables us to jointly solve for both the optimal placement location of the given object in clutter and a corresponding grasp configuration suitable with the placement. We do so using only sensor information of the scene, enabling our approach to work with novel objects. Jointly solving for the grasp and placement configurations ensures compatibility between pick and place by means of propagating gradients. We use a state of the art, grasp learning approach to encode the grasp success likelihood~\cite{Grasp-Hermans-Lu-VoxelInf-RAM-2020}. Like other works using neural networks for learning~\cite{Grasp-RAM7, Grasp-RAM8, Grasp-RAM19, Grasp-RAM22, Grasp-SampleBased-RAM23}, the ability to compute gradients through the model enables efficient gradient-based planning. We detail our proposed solution in Sec.~\ref{sec:approach}.

We evaluate our planner in various placement tasks including,  placing items in a line, tight packing of objects, and object stacking. We define associated costs for each of these tasks, highlighting the modularity of our approach. We validate our approach on a physical robot with a multi-fingered hand. Our results in Sec.~\ref{sec:experiments} show our approach has higher placement success rate than the baselines that treat the individual pick and place planning as sequential, non-interacting problems. Along with improved pick and place success rate, our method is able to handle harder placement configurations with clutter in both pick and place scenes.

A primary limitation to our work as implemented is the assumption that the object maintains the same rigid offset to the hand as that decided in the initial plan. This could easily be incorrect as the object might move during the pick or transit phase. Our work also does not examine any visual or tactile feedback during placement to ensure gentle contact~\cite{romano-tro2011-tactile-skills,sundaralingam-icra2019-tactile-force-learning} with the environment or correct for inaccuracies in planning. Other areas for improvement include placing on sloped or non-planar surfaces. We discuss further ideas for improving our work in Sec.~\ref{sec:conclusion}.

We make the following contributions.
\begin{enumerate}[wide, labelwidth=!, labelindent=0pt]
\item Present a framework for reasoning about pick and place planning jointly. The components of which could be easily swapped to achieve different tasks.
\item Provide a concrete implementation using a learned multi-fingered grasp classifier to encode grasp cost in the objective.
\item Empirically validate the ability of our framework to also pick from clutter.
 \item A fast, GPU-accelerated 3D signed distance generator based on partial view point clouds,  that can be easily reused and updated as the placement scene changes during sequential pick and place executions.
 \item We replicate the grasp learning method of~\cite{Grasp-Hermans-Lu-VoxelInf-RAM-2020} on a different hand, providing further support for its effectiveness.
 \end{enumerate}

\section{Related Work}
\label{sec:related}
We now describe related work in placement and joint pick and place planning. We note that the joint grasp and placement planning problem can be seen as a special case of the more general task-oriented grasping problem~\cite{li-jra1988}. We only examine those task-oriented papers that specifically examine placement.
The robotic object placement problem typically focuses on finding a placement pose for an object, such that it will be stable when released and potentially meet some semantic requirements~\cite{Place-Henrich-GeometricGpuBased2014, Place-DrummondRotationStabilitylearning2021,Place-StoneStack2017,Place-SaxenaPlaceLearn2012}.

Jiang et al. were the first to examine learning for stable object placement of novel objects from partial-view point clouds~\cite{Place-SaxenaPlaceLearn2012}. Their proposed method employs hand-modeled features to learn stable and semantic placement locations for multiple objects in complex scenes and solves the associated inference problem for planning as an integer linear program. Though they have real robot demonstrations, they assume given and feasible grasps.
\cite{Place-Henrich-GeometricGpuBased2014} discusses a GPU based method to generate orientation and contact points for object and environment models constructed from sensors, using local optimization to validate stability. Only placement poses of the object are generated, ignoring any robot constraints.

A few works have examined arbitrary reorientation of objects.
Furrer et al.~\cite{Place-StoneStack2017} show impressive results of stable placement configurations for stacking stones by optimizing over costs generated using a physics simulation.
Newbury et al.~\cite{Place-DrummondRotationStabilitylearning2021} learn stable, human-preferred orientations for placing objects observed as point clouds onto flat surfaces. 

The pick and place method described in~\cite{PnP-CDIM-CoRR2017} learns to pick novel objects in clutter using grasping primitives and drops the objects into bins according to the measured appearance. Zeng et al. then proposed Transporter Networks in \cite{PnP-TransporterNets-Zeng} using spatially consistent visual representation to learn pick-conditioned placing.
Gualtieri et al.~\cite{PnP-AttentionFocusReinfLearning-Gualtieri-Platt-2018} propose a reinforcement learning approach to learn joint pick and place policies trained separately for different object classes.

Haustein et al. propose an anytime algorithm in~\cite{PnP-KragicMonteCarloPlace2019} to solve for optimal and stable placement locations given the object and the environment meshes on user provided heuristic. Though this method is not scalable to novel objects, due to the need for meshes, they propose a similar constrained optimization approach to the one we present.

Zhao et al. propose a task-oriented grasping approach in ~\cite{PnP-Kromer-TaskOrientedGrasp-2020} solving for highly precise task-oriented grasps in $SE(2)$, by filtering sampled grasps through separate networks for predicting grasp quality, post-grasp displacement and task quality, trained using a curriculum learning approach.

Paxton et al.~\cite{Place-SemanticPlacement-Paxton-Hermans-Fox-2021} propose a method to generate stable placement configurations satisfying semantic relationships specified as logical predicates for novel objects. They learn discriminators to predict stability as well as logical predicates and plan for placement locations through a gradient-free optimization.
While this approach uses a learned grasp planner~\cite{sundermeyer-icra2021-contact-graspnet}, grasps are selected after the placement location via rejection sampling, making it inefficient in clutter.

Berscheid et al.~\cite{PnP-SelfSupervisedGraspPlaceEmbedding-Kroger-2020} learn embeddings from top views of 2 fingered grasps and placement images such that feasible grasp and place pairs are close to each other in the latent space. They use this learned space to jointly select grasps and placements; experiments show joint inference of grasp and placement outperforms separate inference in clutter.

Mitash et al.~\cite{PnP-Bekris-TaskDriven-BiManual-Placement} propose a pick and place pipeline with pick, place, handoff (regrasp), and sense actions. They estimate the object geometry in order to generate plans for constrained placement of novel objects. Using a combination of suction and two fingered grasps with simple placement scenes their results show that task oriented grasping and perception perform better than the pick then place methods.

He et al.~\cite{PnP-Pick2Place2023} extend the object reconstruction and grasp planning approach for parallel-jaw grippers from~\cite{Grasp-Nikhil2021}, to placement-aware grasping by learning to generate affordance maps using a NeRF representation of the scene with a reconstructed object SDF. They use a sampling based approach for determining an optimal grasp and placement pair.

Our work builds on the findings that joint pick and place outperforms pick then place planning. In contrast to existing work, we propose a modular pick and place framework for use with multi-fingered grasps on novel objects, that plans over continuous grasps and placement configurations.


\section{Pick and Place as Constrained Optimization}~\label{sec:probdef}
\begin{figure}[h]
	\centering
	\includegraphics[width=0.3\columnwidth]{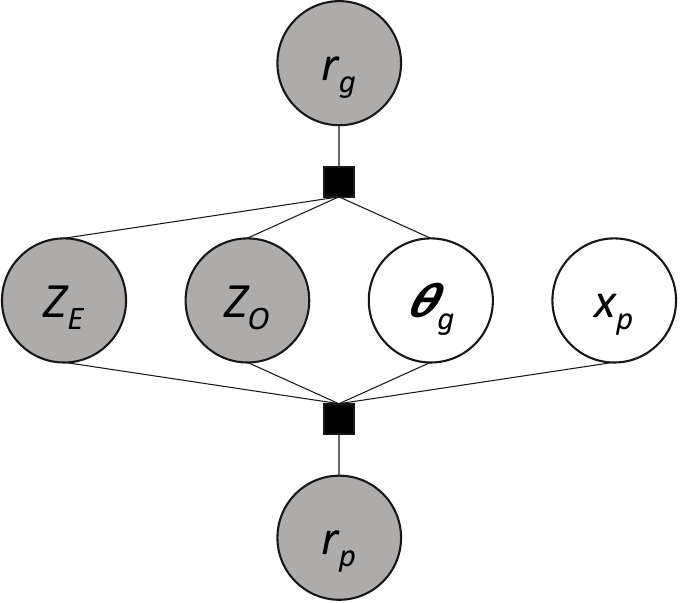}
	\caption{Factor graph of the pick and place probability distribution. We see that while the success probabilities are conditionally independent given the planning parameters, they can not fully decouple, requiring joint inference over pick and place parameters.\label{fig:factor-graph}}
\end{figure}

Let $O$ be an object to be placed in a cluttered environment $E$, with partial-view depth images $Z_O$ and $Z_E$ respectively. The grasp configuration $\thetav_{g} = [\xv_g, \qv_g^h]$ is a vector including the robot palm pose $\xv_g \in SE(3)$ and preshape joint angles of the gripper's fingers $\qv_g^h \in \mathcal{Q}_h$. The placement configuration $\xv_p \in SE(3)$ defines the 6-DOF pose of where the centroid of object point cloud $Z_O$ should be once placed. We can then define the probability of successfully grasping the object as:
\vspace{-0.5em}
\begin{equation}
 P\left(r_g{=}1|\thetav_{g},Z_O\right) = F\left(\thetav_{g}; Z_O\right) \label{eq:gprob} \vspace{-0.5em}
\end{equation}
and the probability of the place configuration $x_p$ being successful for object $O$ in environment $E$ as:
\vspace{-0.5em}
\begin{equation}
 P\left(r_p{=}1|\xv_p, \thetav_g,Z_O, Z_E\right) = G\left(\xv_p; \thetav_g, Z_O, Z_E\right) \label{eq:pprob}  \vspace{-0.5em}
\end{equation}
The joint probability for pick and place success is then:
\vspace{-0.5em}
\begin{multline}
  P\left(r_g{=}1, r_p{=}1|\thetav_{g}, \xv_p, Z_O, Z_E\right)\\ = F\left(\thetav_{g}; Z_O\right)G\left(\xv_p; \thetav_g, Z_O, Z_E\right) \label{eq:jointprob}  \vspace{-0.5em}
\end{multline}
Which we visualize as a factor graph in Fig.~(\ref{fig:factor-graph}). We see that while the success probabilities are conditionally independent given the planning parameters, they do not fully decouple, requiring joint inference over pick and place parameters.

We define the pick and place inference problem as finding a tuple of grasp configuration and place configurations $(\thetav_{g}, \xv_p)$, that maximizes the joint probability defined in Eq.~(\ref{eq:jointprob}). Taking the negative log on Eq.~(\ref{eq:jointprob}), we formalize this as a constrained optimization in Eq.~(\ref{eq:opt})
\vspace{-2pt}
\begin{subequations}
 \label{eq:opt}
 \begin{flalign}
\argmin_{\xv_p, \thetav_g, \qv_g^a, \qv_p^a, \tau} &{-}{\ln}{\left(G\left(\xv_p; \thetav_g, Z_O^+, Z_E\right)\right) -}{\ln}{\left(F\left(\thetav_g; Z_O\right)\right)}\label{eq:opt_cost}\\
 \text{subject to} \quad \xv_p & \in \mathcal{P} \label{eq:placement_set} \\
 \xv_g &= \phi_h(\qv_g^a); \quad \xv_p = \phi_O(\qv_p^a) \label{eq:arm_fk} \\
 \qv_i^- &\leq \qv_i \leq \qv_i^+ \quad \forall \; i \in \{g,p\} \label{eq:CBoundG} \\
 Z_O^+ &= Z_O \cup R_{G}(\thetav_g) \label{eq:object_cloud_augmentation} \\
 \epsilon &\leq \sdf\left(\xv_p,Z^+_O\left(\thetav_g\right),Z_E\right) \label{eq:CColl1} \\
 \tau(\xv_p, \xv_g) &\in \Omega \label{eq:CTraj}
 \end{flalign}
\end{subequations}
Equation~(\ref{eq:opt_cost}) defines the objective of the optimization as a log-linear combination of the placement success probability $G\left(\xv_p; \thetav_g, Z_O, Z_E\right)$ and grasp success probability $F\left(\thetav_g; Z_O\right)$.
The remaining constraints ensure physical validity for successful execution, i.e., the grasp and placement must be reachable by the robot and the objects and robot should not interpenetrate.
Eq.~(\ref{eq:placement_set}) constrains the placement configurations $\xv_p$ to be within the footprint of the placement surface and above it. Depending on the task, the placement configuration $\xv_p$ is in $SE(3)$ or $SE(2)$.
Eq.~(\ref{eq:arm_fk}) encodes the arm forward kinematics for the grasp and placement, while Eq.~(\ref{eq:CBoundG}) defines the joint limits, where the superscript, \(i\), denotes the joints associated with the robot arm and hand.

\begin{figure}
	\centering
	\begin{subfigure}{0.32\columnwidth}
		\phantomsubcaption \label{fig:aug_obj}
	\end{subfigure}
	\begin{subfigure}{0.32\columnwidth}
		\phantomsubcaption \label{fig:aug_robot}
	\end{subfigure}
	\begin{subfigure}{0.32\columnwidth}
		\phantomsubcaption \label{fig:aug_full}
	\end{subfigure}
	\includegraphics[width=1\columnwidth, trim={0 170 0 160}, clip]{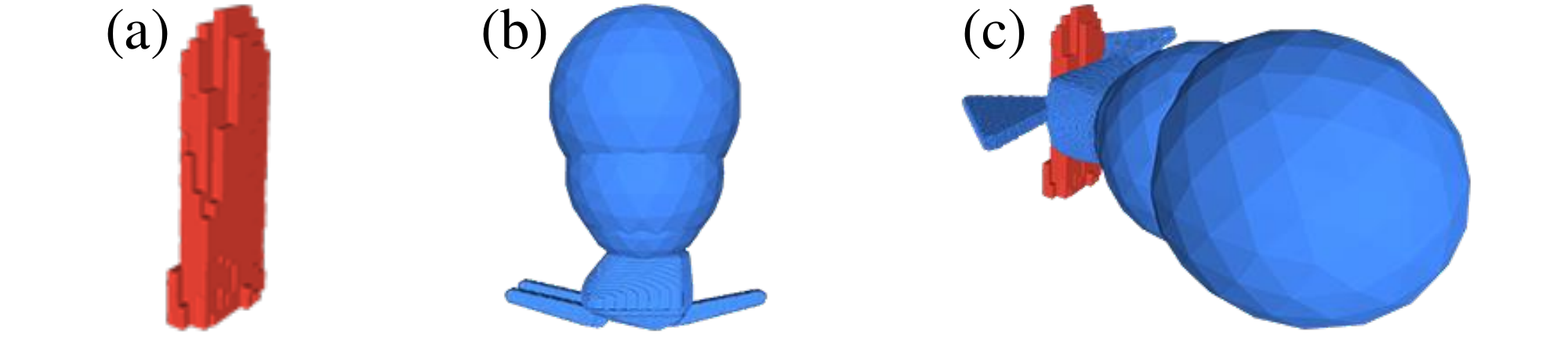}
	\caption{Robot-object geometry for pick-and-place collision checking: (\subref{fig:aug_obj}) object; (\subref{fig:aug_robot}) robot; (\subref{fig:aug_full}) union of object and robot. \label{fig:aug}}
	\vspace{-12pt}
\end{figure}

Eq.~(\ref{eq:object_cloud_augmentation}) augments the object cloud with the geometry of the hand and spheres approximating the geometry of the wrist (last 2 arm links) defined by $R_{G}(\thetav_g)$, at the current grasp pose. We note this is a similar procedure to that in~\cite{Packing-RobotPackingHauser19}. We visualize the gripper geometry augmentation in Fig.~(\ref{fig:aug}).
Using this we define the placement collision constraint in  Eq.~(\ref{eq:CColl1}).
Finally, Eq.~(\ref{eq:CTraj}) defines that there must be a feasible, collision-free trajectory from grasp to placement.

\section{Solving the Joint Pick and Place Problem}\label{sec:approach}
In this section, we discuss our approach to instantiating and solving the problem defined by Eq.~(\ref{eq:opt}). We first discuss the details of different placement probabilities, \(G(\cdot)\), which we examine in our experiments. We then briefly review the learning-based grasp method from~\cite{Grasp-Hermans-Lu-VoxelInf-RAM-2020} and its use as our grasp probability \(F(\cdot)\). Following that we present an efficient algorithm for SDF-based collision checking built specifically for repeated placement into clutter. We conclude this section by discussing the choice of solver used and the generation of grasp and place priors to perform MAP inference.

\subsection{Placement Probabilities} \label{sec:place_heuristics}
Given any placement cost $H(\xv_p)$ that accepts an object placement configuration $\xv_p \in SE(3)$ as input and outputs a scalar value that quantifies the suitability of $\xv_p$ for the task considered, with lower values being more desirable for the task and higher values being less desirable.

We can convert this $H(\xv_p)$ to a probability likelihood as:
\vspace{-3pt}
\begin{equation}
	G(\xv_p) \propto \exp{\left(-\alpha H(\xv_p)\right)} \label{eq:pheur} \vspace{-0.5em}
\end{equation}
Higher values of the $\alpha$ parameter make the solver prefer placements more desirable to the task at the expense of grasp success, lower values of $\alpha$ prefer more confident grasps.

We now define the four placement likelihoods used in this paper.
 We use the notation $G(\xv_p)$ to denote placement likelihood and $H(\xv_p)$ to denote placement cost. Any cost we define can be converted to a likelihood using Eq.~(\ref{eq:pheur}).

\subsubsection{Target Pose} The simplest cost that encodes placing the object to a target pose. We can define this as simply minimizing the squared-Euclidean distance between the selected and target pose, giving the likelihood:
\begin{equation}
	\begin{split}
		G_\text{target}(\xv_p;\xv_t) \propto \exp{\left(-\frac{1}{2}\left(\xv_t - \xv_p\right)^{T}\left(\xv_t - \xv_p\right)\right)}
	\end{split} \label{eq:h_target}
\end{equation}

\subsubsection{Tight Packing} The cost defined in Eq.~(\ref{eq:h_pack}) aims to place objects as close to each other as possible. This is relevant for organizing objects into shelves or boxes. We encode this cost as the area of the bounding box enclosing all objects in the scene plus the area of the bounding box enclosing the newly placed object and a reference point.
  \begin{align}
    H_\text{pack}(\xv_p;Z_O, Z_E) &= L_E(\xv_p, Z_O, Z_e) \cdot W_E(\xv_p, Z_O, Z_e) \notag \\
& + (1, 1, 0)\cdot T_2(\xv_p)\cdot (L_O, W_O, 1)^T \label{eq:h_pack}
  \end{align}
 where $(L_E, W_E)$ define the length and width of the bounding box enclosing all objects in $Z_E$, including the newly placed object, $(L_O, W_O)$ define the length and width of the object point cloud $Z_O$, and \(T_2(\xv_p)\) defines the homogeneous $SE(2)$ transformation matrix associated with pose \(\xv_p\), the place probability for this cost is obtained using Eq.~(\ref{eq:pheur}). Fig.~(\ref{fig:pack_exec}) shows results of planning with the tight packing cost.
\subsubsection{Stacking} Allows objects to be placed on top of each other. Given the centroid $x_c$ of the point cloud of an existing object or stack $Z_E$, the cost defined in Eq.~(\ref{eq:h_stack}) penalizes the height $H_O$ and width $L_O$ of the object at place configuration $x_p$, to lower the height of the stack and improve object alignment for stability. The position of the object is constrained to be close to $x_c$ while the orientations are not constrained. This task allows for placement configurations $\xv_p$ to be in $SE(3)$.
Fig.~(\ref{fig:demo_stack}) shows the robot stacking five blocks using this cost.
\begin{equation}
	\begin{split}
		H_\text{stack}(\xv_p; Z_O, Z_E) = (1, 0, 1) \cdot \hat{R}(\xv_p) {\cdot} (L_O, W_O, H_O)^T
	\end{split} \label{eq:h_stack}
\end{equation}

where $\hat{R}(\xv_p) = \hat{R}_x(\theta)\hat{R}_y(\psi)\hat{R}_z(\phi)$ and $\hat{R}_k$ defines the absolute values of the rotation matrix about object-axes \(k\).

\subsubsection{Place Inline} Places a sequence of objects in a straight line given a point on the line $x_t$ and its angle of slope $\theta_l$. We model this as a Gaussian about the \(x\) component of the placement configuration in Eq.~(\ref{eq:h_inline}).
\begin{equation}
	G_\text{inline}(\xv_p; x_t, \theta_l) = -\exp\left({-(\xv_p - x_t)^TK_{\theta_l}(\xv_p - x_t)}\right) \label{eq:h_inline}
\end{equation}
where $K_{\theta_l} = R_z(\theta_l)^TK_xR_z(\theta_l)$ with $R_z(\theta_l) \in SO(2)$
and $K_x = \left[\begin{array}{cc}1 & 0\\ 0 & 0 \end{array}\right]$, aligns the \(x\)-coordinates.
 Fig.~(\ref{fig:demo_inline}) shows the real robot rearranging a set of cups in a straight line.
 \subsection{Grasp Prediction}

\begin{figure}
	\vspace{3pt}
	\centering
	\includegraphics[width=\columnwidth, trim={100 80 100 70}, clip]{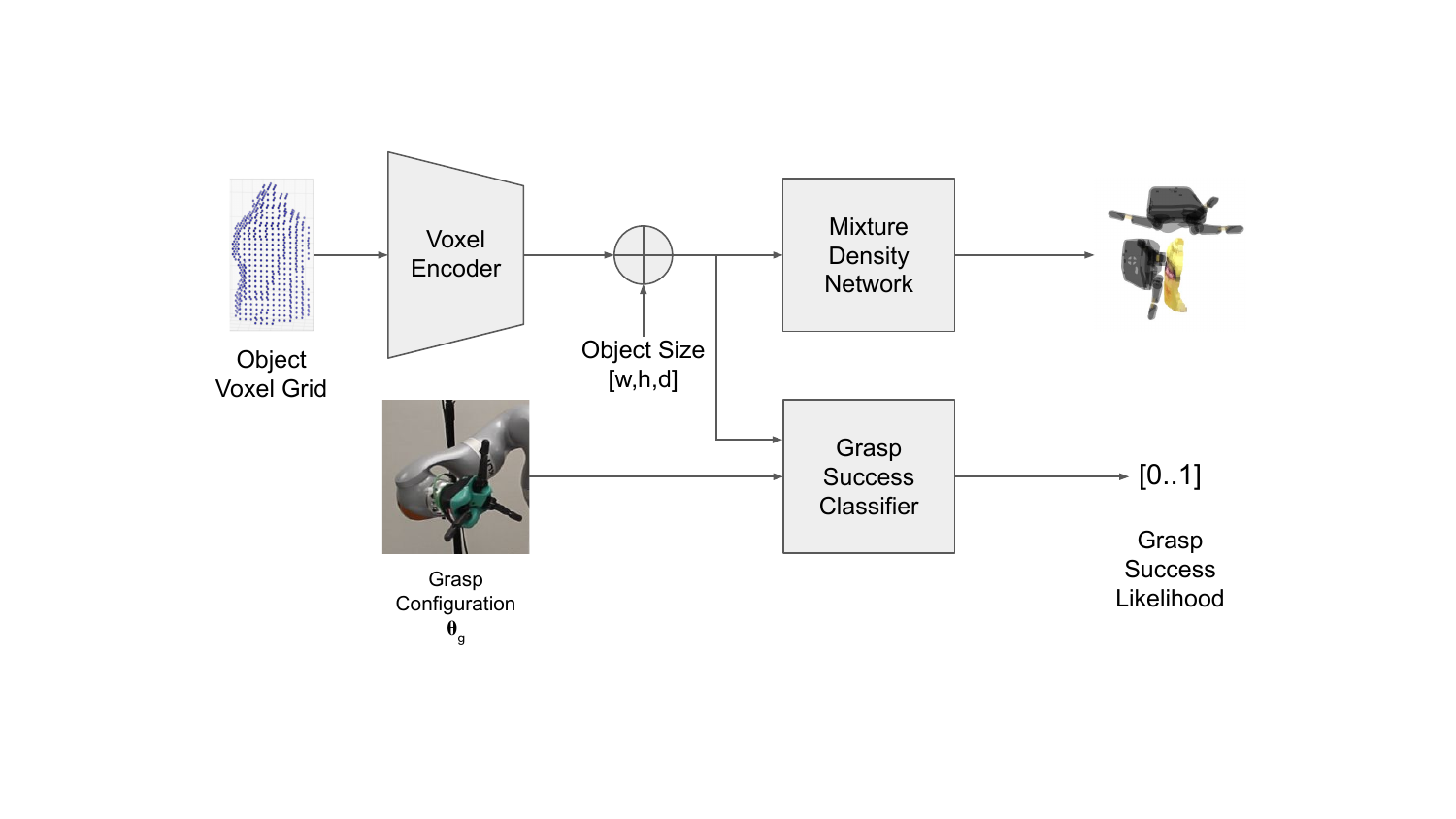}
	\caption{Overview of our grasp prediction pipeline based on~\cite{Grasp-Hermans-Lu-VoxelInf-RAM-2020}. A grasp classifier predicts grasp success given an object voxel grid and grasp configuration. A mixture density network models a  distribution over grasp configurations given an input voxel grid.}
	\label{fig:grasp_net}
	\vspace{-15pt}
\end{figure}
 
Following the success of recent learning-based grasp planning approaches~\cite{Grasp-Hermans-Lu-VoxelInf-RAM-2020,Grasp-Fox-6DOF-Grasp-Net-2019} we define our grasp cost  as the negative log probability of grasp success $-\log{\left(F(\thetav_g; Z_O)\right)}$. However, we note that our framework could handle any differentiable grasp cost encoding.

Fig.~(\ref{fig:grasp_net}) shows an overview of the grasp prediction pipeline used here. A neural network classifier defines the core of the grasp prediction model $F(\thetav_g; Z_O)$. This outputs a scalar value between 0 and 1 that represents the grasp success probability for the given grasp $\thetav_g$ on the observed object $Z_O$. We learn this model as a 3D convolutional neural network classifier using the approach proposed in~\cite{Grasp-Hermans-Lu-VoxelInf-RAM-2020}. This takes a voxel representation of the object, converted from the point cloud, as input and passes it through several 3D convolutional layers to predict grasp success. The only modification we make to the neural network structure is changing the grasp input model to accept the one-dimensional preshape configuration (the finger spread) instead of the higher-dimensional vector used for the dexterous hand in~\cite{Grasp-Hermans-Lu-VoxelInf-RAM-2020}.

\label{sec:mdn}A mixture density network (MDN) comprises an additional part of the model. It takes the same voxel grid as input and generates a distribution over grasp configurations as output (c.f. Fig.~(\ref{fig:mdn})). We use this to initialize the solver described later in this section. For further details see~\cite{Grasp-Hermans-Lu-VoxelInf-RAM-2020}.

\subsection{SDF Collision Constraint Computation}\label{sec:sdf}

\begin{figure}
	\centering
	\begin{subfigure}[b]{0.32\columnwidth}
		\phantomsubcaption
		\label{fig:coll-dilator}
	\end{subfigure}
	\begin{subfigure}[b]{0.32\columnwidth}
		\phantomsubcaption
		\label{fig:coll-sdf_list}
	\end{subfigure}
	\begin{subfigure}[b]{0.32\columnwidth}
		\phantomsubcaption
		\label{fig:coll-full_sdf}
	\end{subfigure}
	\begin{subfigure}{0.32\columnwidth}
		\phantomsubcaption
		\label{fig:coll-grasp_binary}
	\end{subfigure}
	\begin{subfigure}{0.32\columnwidth}
		\phantomsubcaption
		\label{fig:coll-place_binary}
	\end{subfigure}
	\begin{subfigure}{0.32\columnwidth}
		\phantomsubcaption
		\label{fig:coll-place_init}
	\end{subfigure}
	\includegraphics[width=1\columnwidth]{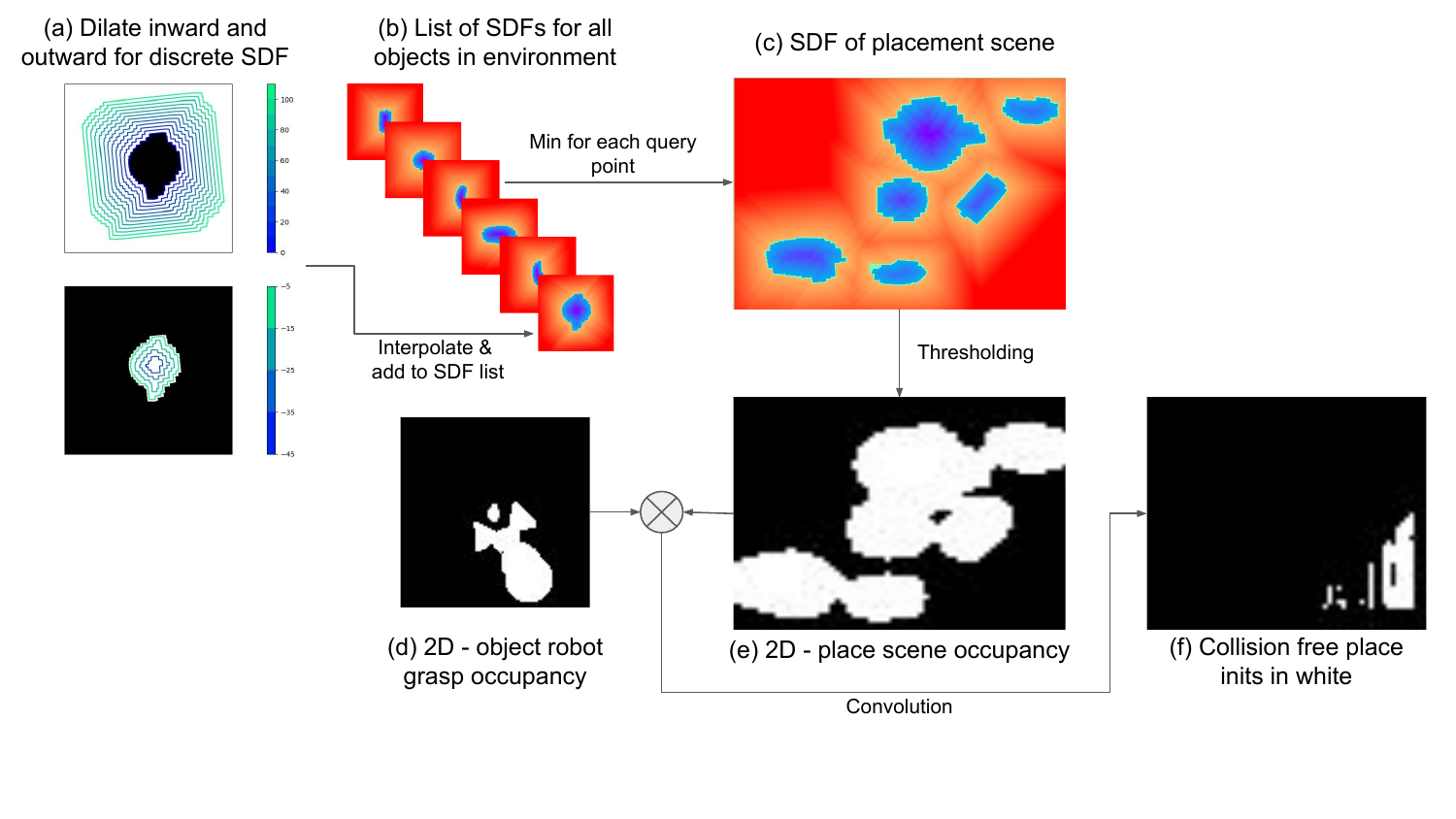}\vspace{-12pt}
	\caption{Steps to generate signed-distance function for place scene and computing collision-free place initializations.}
	\label{fig:coll}
\end{figure}

To account for the collision constraints in Eq~(\ref{eq:CColl1}), we require signed distances from the partial view points of the objects in the environment. For efficient computation we compute a discrete approximation of the SDF that we can quickly update as more objects are placed into the scene.
Fig.~(\ref{fig:coll}) shows the steps in generating the SDF queries for the collision constraint.

Given a point cloud of the environment we convert it to a 3D voxel grid encoding the point cloud occupancy.
We then interpret this occupancy grid  as a discretization of the zero level set of the environment's SDF.
To compute the positive signed distances associated with the surface we use a brushfire algorithm to iteratively march outward till a truncation distance from the zero level set to obtain discrete positive distances at uniform increments.
Similarly the negative signed distances are obtained by marching inward, resulting in a truncated discretized signed distance field (\dsdf{}).
By treating the \dsdf{} as a 3D image we apply 1D finite differentiation filters along the \(x\), \(y\) and \(z\)-axes to compute the gradients.

Given the discrete SDF, we can query it with any continuous point with trilinear interpolation of neighboring points. To handle the collision constraints in the optimization problem Eq.~(\ref{eq:opt}), we obtain a uniform discretized set of points associated with the object being placed, $Z_O$, and the known robot geometry, $R(\thetav_g)$ denoted as \(\mathbb{Z}_{o}\) and \(\mathbb{Z}_R(\thetav_g)\) respectively, where we keep the dependence on the grasp configuration explicit.
For use in the optimization we transform these points $\mathbb{Z} = \mathbb{Z}_{o} \cup \mathbb{Z}_R(\thetav_g)$ according to the placement pose, \(\xv_p\), this transformation and querying is done  efficiently in parallel as defined below:
\begin{equation}
	\sdf\left(\xv_p, \mathbb{Z}, {Z_E}\right) = \min_{\xv \in \mathbb{Z}, \mathbb{Z} \in \{\mathbb{Z}_O, \mathbb{Z}_R\}} \dsdf_E\left(T({\xv_p})\mathbb{Z}\right) \label{eq:query_sdf}
      \end{equation}
 where the min over both object and robot geometry accounts for the union operation Eq.~(\ref{eq:object_cloud_augmentation}). In practice we can treat the collision constraints for the object $O$ and the robot $R$ separately for the same grasp and place configurations to have more informative gradients.


\subsection{Optimization and Motion Planning}
\begin{figure}
	\begin{subfigure}[b]{0.24\columnwidth}
		\phantomsubcaption \label{fig:mdn_lego}
	\end{subfigure}
	\begin{subfigure}[b]{0.24\columnwidth}
		\phantomsubcaption \label{fig:mdn_cracker}
	\end{subfigure}
	\begin{subfigure}[b]{0.24\columnwidth}
		\phantomsubcaption \label{fig:mdn_mustard}
	\end{subfigure}
	\begin{subfigure}[b]{0.24\columnwidth}
		\phantomsubcaption \label{fig:mdn_pitcher}
	\end{subfigure}
  \includegraphics[width=\columnwidth, trim={65 180 50 180}, clip]{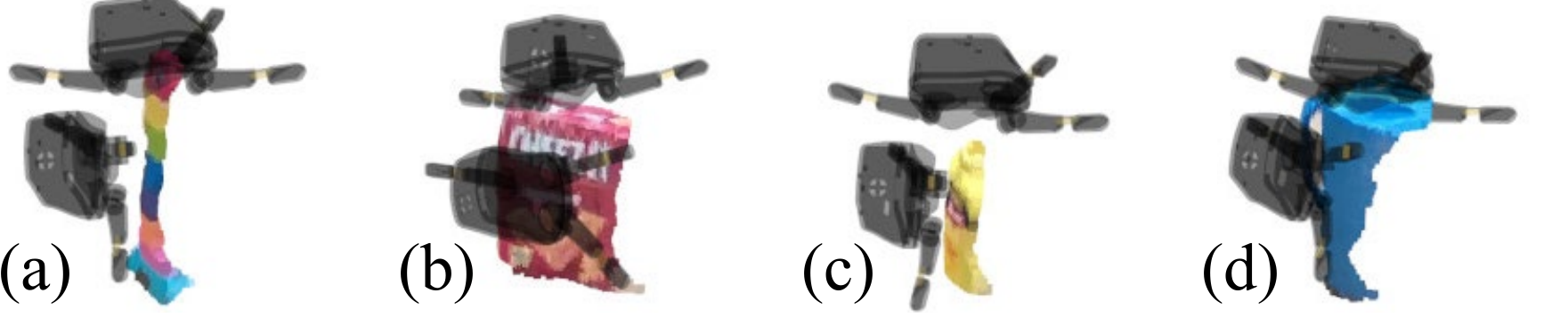}
  \caption{Mean configurations of the MDN top and side grasp modes, visualized with the partial view point clouds of the objects (\subref{fig:mdn_lego}) lego blocks, (\subref{fig:mdn_cracker}) cracker box, (\subref{fig:mdn_mustard}) mustard bottle and (\subref{fig:mdn_pitcher}) pitcher.}
  \label{fig:mdn}
  \vspace{-15pt}
\end{figure}
We perform MAP inference by solving the optimization problem from Eq.~(\ref{eq:opt}) without constraint Eq.~(\ref{eq:CTraj}). We relax the forward kinematics Eq.~(\ref{eq:arm_fk}) and  the collision SDF constraints Eq.~(\ref{eq:CColl1}) into the objective using an Augmented Lagrangian method.
We convert the constraint, \(\xv \in \mathcal{P}\), to bound constrain the placement configuration, \(\xv_p\), within the table edges.
We solve the resulting bound constrained problem using BFGS~\cite{nocedal2006numerical} with projections to handle the bounds on the joint angles and placement pose. We ensure Eq.~(\ref{eq:CTraj}) when motion planning for the arm after solving for the pick and place configurations.

The solver is initialized with grasp configurations ${\thetav_g}^0$ sampled from the MDN prior described in \ref{sec:mdn} and shown in Fig.~(\ref{fig:mdn}). The place prior is then obtained by convolving the 2D binary occupancy of the augmented object-robot geometry (Fig.~( \ref{fig:coll-grasp_binary})) over the coarse 2D binary occupancy of the place scene (Fig.~( \ref{fig:coll-place_binary})), which outputs collision-free place configurations (Fig.~( \ref{fig:coll-place_init})), these are then ranked by predicted place probabilities and filtered by kinematic feasibility, to obtain the initial place configuration ${\xv_p}^0$.

Given the solutions of the grasp and place configurations in joint space, the full mesh models of the robot, the grasp and placement surfaces, and meshes of all objects in the scene generated from the computed SDFs , we use MoveIt!'s constrained RRT planner~\cite{moveit} for planning trajectories to pre-grasp pose, transfer from pick to place configurations, and pre-place pose. We constrain the task space orientation of the hand to restrict rotation of the object being during transfer to the place pose. The pre-grasp and pre-place poses are obtained by offsetting the end-effector by a small distance along the negative direction of the end-effector's x-axis. We use a task space velocity controller during grasp approach, lift, place approach, and post-placement retraction to soften interaction with the environment.

\vspace{-5pt}
\section{Experiments} \label{sec:experiments}
\vspace{-3pt}
We now validate the benefits of our proposed pick and place framework (i.) by benchmarking against sequential pick and place and a sampling based baseline, and (ii.) qualitative experiments demonstrating the applicability of the approach to a variety of scenarios.
We experiment using a KUKA iiwa 7-DOF arm with a Reflex Takktile 2 gripper and a Realsense D455 camera for sensing,  with an 8-core Ryzen 5800X CPU and a 12GB Nvidia RTX 3060 for computation.
\subsection{Benchmarks}
We benchmark the success rate of pick and place executions and the optimality of our joint optimization approach against the baselines in 2 different tasks, the following baselines are considered:

\begin{enumerate}[wide, labelwidth=!, labelindent=0pt]
 \item \textbf{The pick then place approach:} Where we solve for the best grasp configuration subject to all the constraints in Eq.~(\ref{eq:opt}) that apply during grasping. Then keeping this grasp configuration fixed,  the placement configuration that suits the grasp is solved for. This is essentially done by solving the optimization problem described in \ref{sec:probdef} and \ref{sec:approach} twice with the grasp and place costs individually. This is similar to the pick-conditioned placing in \cite{PnP-TransporterNets-Zeng} with grasps in $SE(3)$.
 \item \textbf{Sampling:} Similar to the approach in \cite{Place-SemanticPlacement-Paxton-Hermans-Fox-2021, PnP-Pick2Place2023},  We develop a baseline that generates compatible grasp and placement configurations using Monte Carlo sampling. First we sample a set of grasp configurations with high success rate from the trained grasp MDN network, and a set of placement configurations not in collision with the environment for both object and robot using the	generated SDF, then the generated grasp and placement configurations are refined locally for feasibility with other constraints.
\end{enumerate}


\subsubsection{Place to an unreachable target}

The robot is tasked with picking an object in isolation and placing it on a corner of a table (placement surface) that is not reachable by the robot. We recorded 30 executions for this task with 10 different objects in varied levels of clutter ranging from 4 -- 7 objects in the placement scene. We use the target pose cost defined in Eq.~(\ref{eq:h_target}). Since, the target pose is unreachable and may be infeasible due to other objects being in the way, the solver must find a feasible placement solution as close to the target as possible while accounting for the grasp.

\begin{figure}
	\vspace{3pt}
	\centering
	\includegraphics[width=\columnwidth]{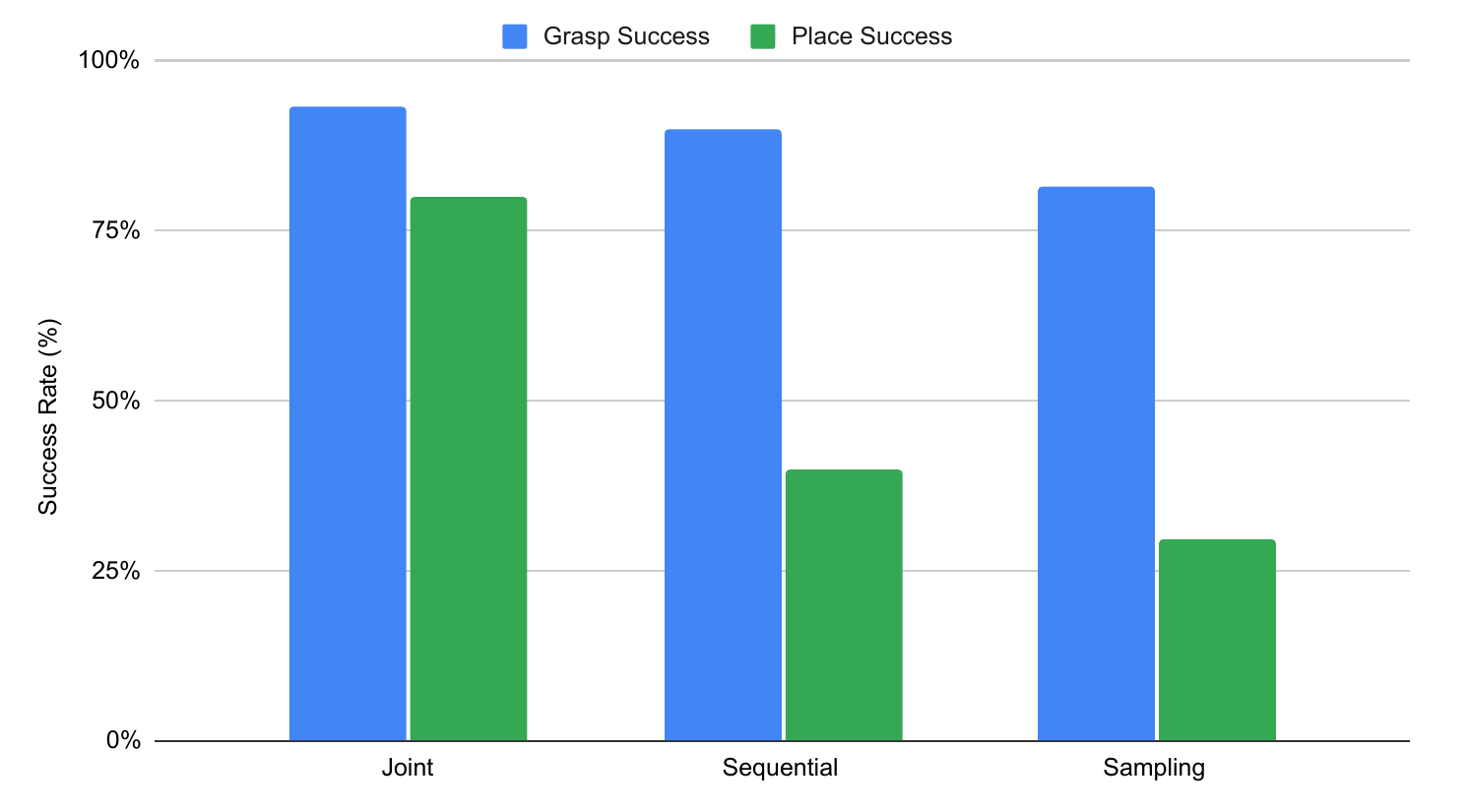}\vspace{-0.5em}
	\caption{Grasp and place success rates for the joint inference, sequential inference and sampling methods across 30 executions, for placing into clutter benchmark.}
	\label{fig:benchmark_suc}
	\vspace{-12pt}
\end{figure}

We report the success rate, and predicted placement probabilities for each method. Fig.~(\ref{fig:benchmark_suc}) shows the grasp and place success rates.
We call a grasp successful if the robot lifts the object without dropping it.
  We call a placement successful if the robot places it on the placement surface and all objects remain upright.
  Cases where the solver fails to find a feasible solution are failures. For this task the joint method has an average run time of 58.21s with a standard deviation of 26.83s for varied levels of clutter, the sequential method takes on average 71.48s (37.27s std dev); sampling takes on average 39.93s (5.63s std dev) for 450 samples.

We see that the joint method significantly outperforms the baselines in terms of place success with 80\% of the executions being successful, While unsurprisingly the joint and sequential methods achieve comparable grasp success rates.
The sampling based baseline is not able to reliably generate safe and stable placement configurations. It was often in violation of constraints, due to the initial discretization of the placement configuration samples and lack of gradient information for refinement. The sequential baseline primarily fails due to the lack of feasible placement initializations for the fixed grasp generated. Another interesting cause of failure was the object slipping and falling during the placement trajectory more than the joint optimization method, we hypothesize this could be due to the joint method preferring more tighter grasps to avoid collisions with other objects.

\begin{figure}
	\centering
	\includegraphics[width=\columnwidth, trim={20 0 20 45}, clip]{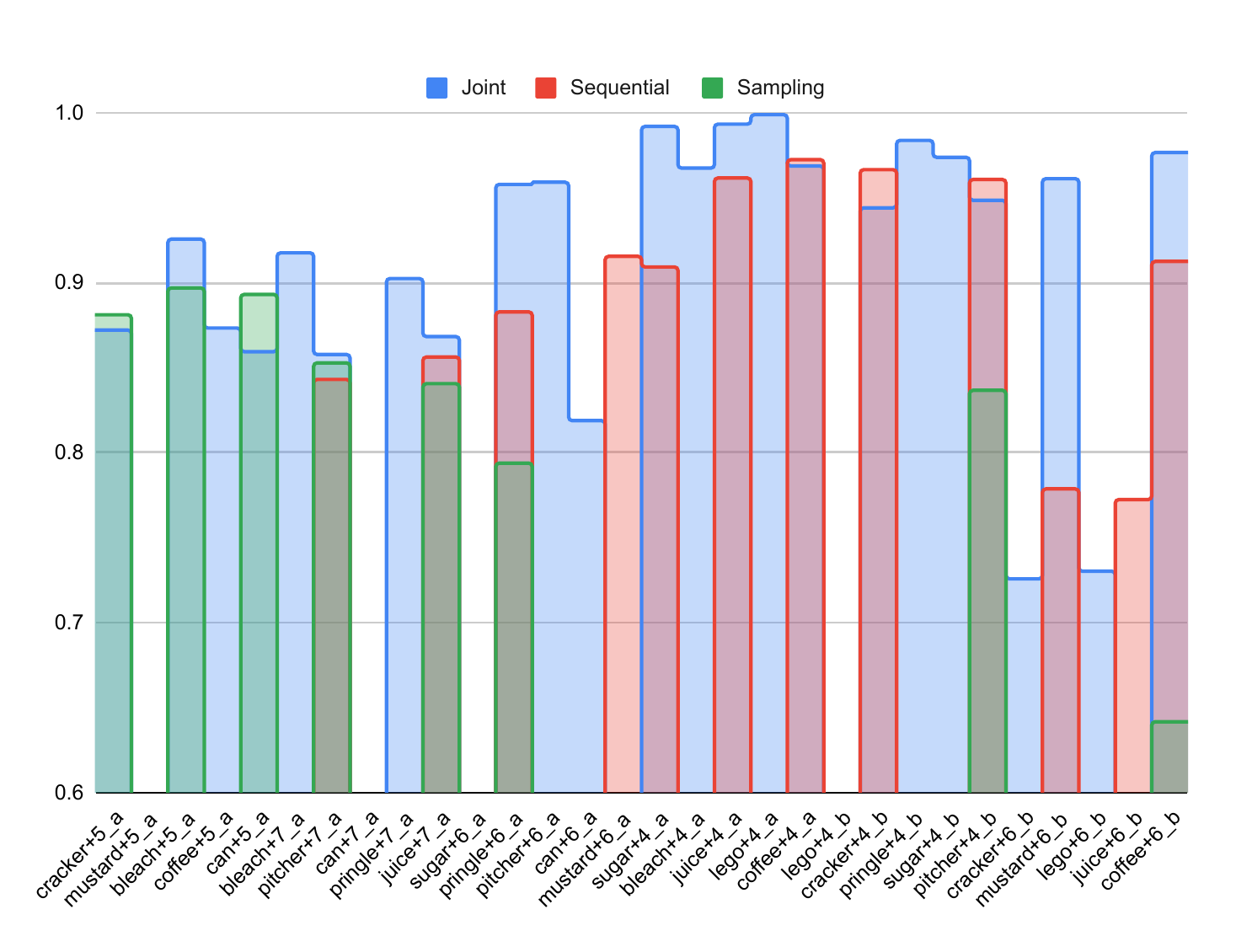}\vspace{-0.6em}
	\caption{Closeness of solved placement configuration to target pose as likelihood $[0,1]$. (1 being the closest and 0 being farthest away). Likelihood of failed executions are set to zero.} \label{fig:benchmark_cost}
	\vspace{-14pt}
\end{figure}

\begin{figure}
	\centering
	\begin{subfigure}{0.31\columnwidth}
		\includegraphics[width=\columnwidth, trim={40 0 40 0}, clip, cfbox=green 1pt 0pt]{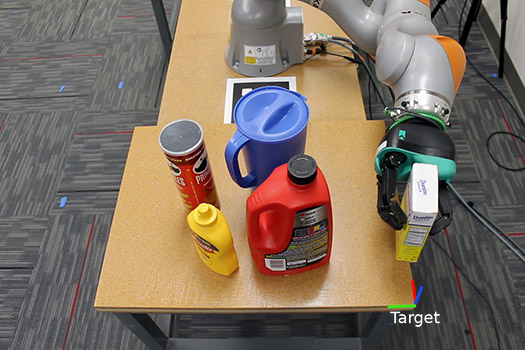}\\[0.5mm]
		\includegraphics[width=\columnwidth, trim={40 0 40 0}, clip, cfbox=green 1pt 0pt]{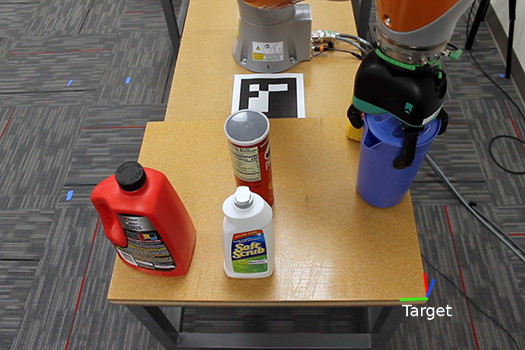}\\[0.5mm]
		\includegraphics[width=\columnwidth, trim={40 0 40 0}, clip, cfbox=green 1pt 0pt]{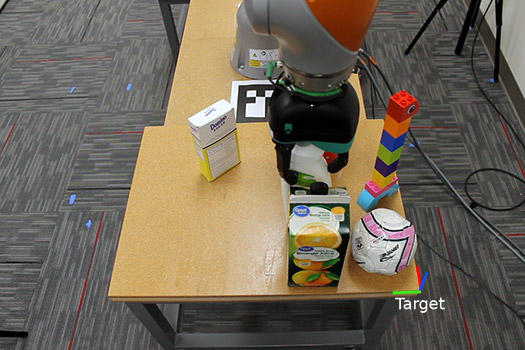}\\[0.5mm]
		\includegraphics[width=\columnwidth, trim={40 0 40 0}, clip, cfbox=green 1pt 0pt]{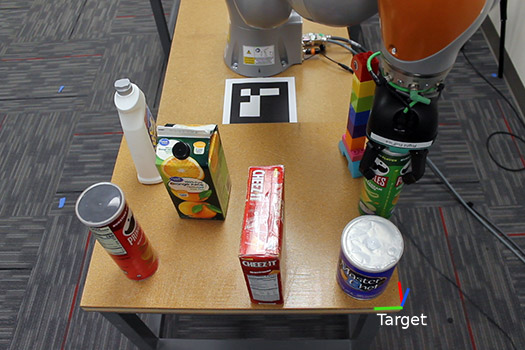}\\[0.5mm]
		\includegraphics[width=\columnwidth, trim={40 0 40 0}, clip, cfbox=green 1pt 0pt]{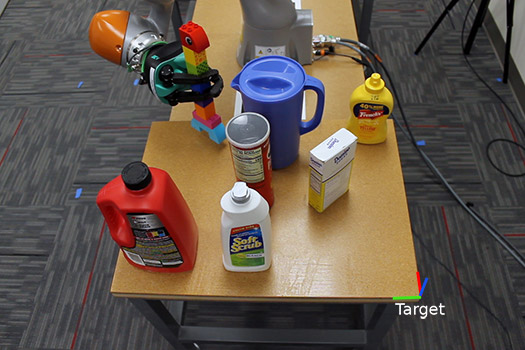}\\[0.5mm]
		\includegraphics[width=\columnwidth, trim={40 0 40 0}, clip, cfbox=green 1pt 0pt]{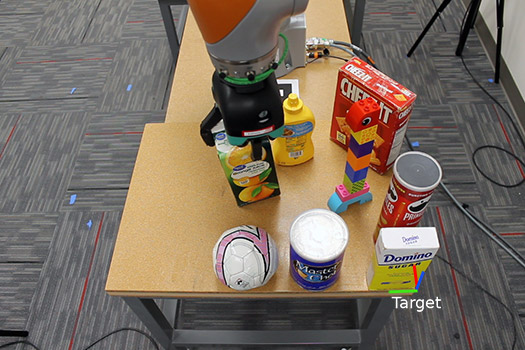}\vspace{-0.5em}
		\caption{}\label{fig:exec-joint}
	\end{subfigure}\hspace{0.02\columnwidth}
	\begin{subfigure}{0.31\columnwidth}
		\includegraphics[width=\columnwidth, trim={40 0 40 0}, clip, cfbox=green 1pt 0pt]{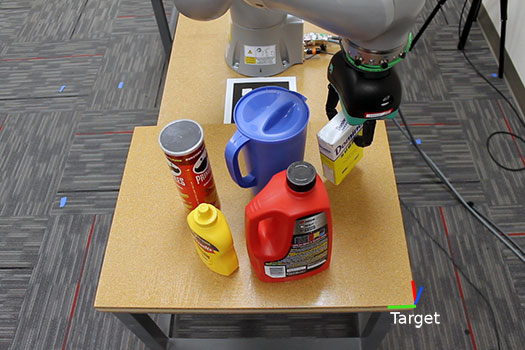}\\[0.5mm]
		\includegraphics[width=\columnwidth, trim={40 0 40 0}, clip, cfbox=green 1pt 0pt]{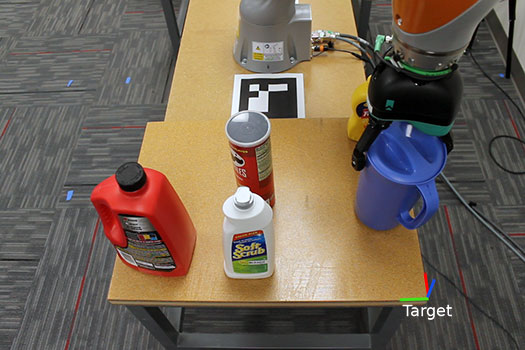}\\[0.5mm]
		\includegraphics[width=\columnwidth, trim={40 0 40 0}, clip, cfbox=red 1pt 0pt]{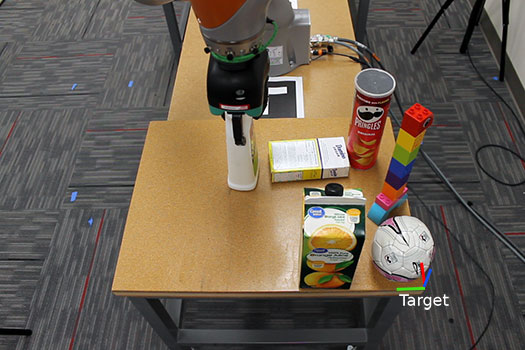}\\[0.5mm]
		\includegraphics[width=\columnwidth, trim={40 0 40 0}, clip, cfbox=green 1pt 0pt]{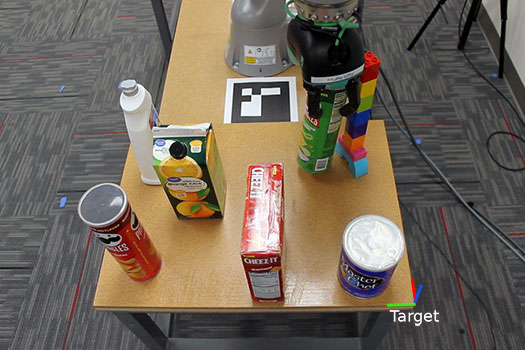}\\[0.5mm]
		\includegraphics[width=\columnwidth, trim={40 0 40 0}, clip, cfbox=red 1pt 0pt]{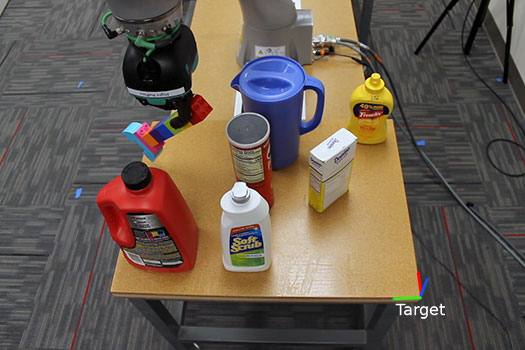}\\[0.5mm]
		\includegraphics[width=\columnwidth, trim={40 0 40 0}, clip, cfbox=green 1pt 0pt]{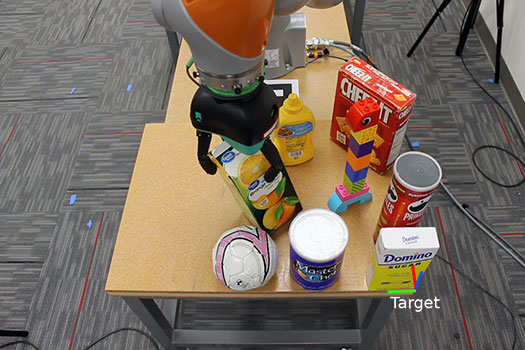}\vspace{-0.5em}
		\caption{}\label{fig:exec-then}
	\end{subfigure}\hspace{0.02\columnwidth}
	\begin{subfigure}{0.31\columnwidth}
		\includegraphics[width=\columnwidth, trim={40 0 40 0}, clip, cfbox=red 1pt 0pt]{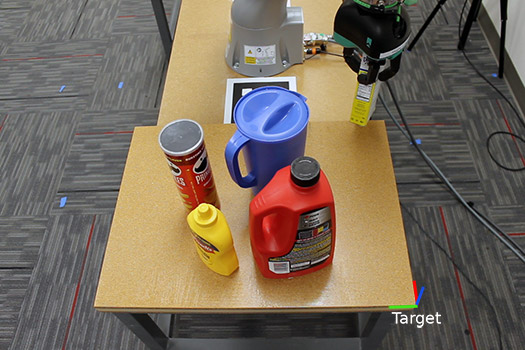}\\[0.5mm]
		\includegraphics[width=\columnwidth, trim={40 0 40 0}, clip, cfbox=green 1pt 0pt]{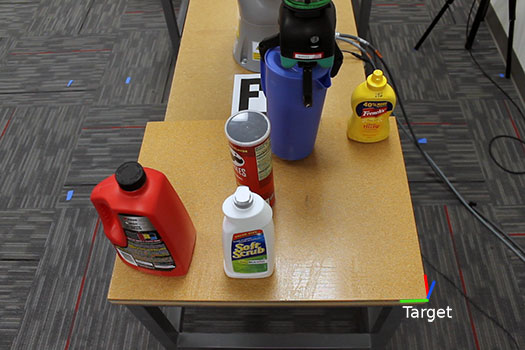}\\[0.5mm]
		\includegraphics[width=\columnwidth, trim={40 0 40 0}, clip, cfbox=green 1pt 0pt]{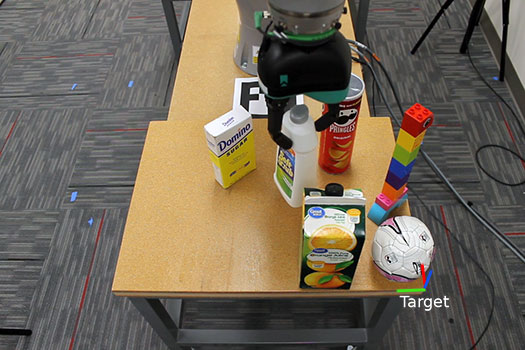}\\[0.5mm]
		\includegraphics[width=\columnwidth, trim={40 0 40 0}, clip, cfbox=green 1pt 0pt]{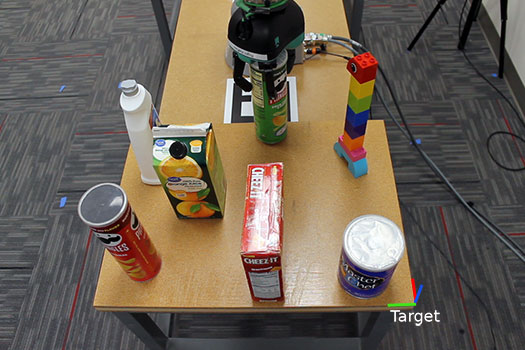}\\[0.5mm]
		\includegraphics[width=\columnwidth, trim={40 0 40 0}, clip, cfbox=red 1pt 0pt]{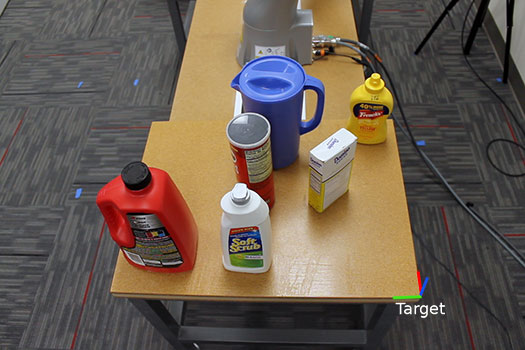}\\[0.5mm]
		\includegraphics[width=\columnwidth, trim={40 0 40 0}, clip, cfbox=green 1pt 0pt]{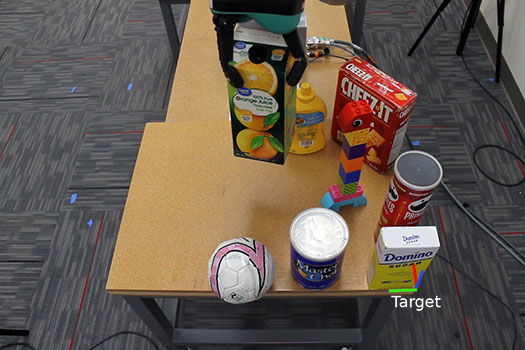}\vspace{-0.5em}
		\caption{}\label{fig:exec-sampling}
	\end{subfigure}

	\caption{Example executions for each scene in the unreachable target task: \textbf{(a)} joint inference, \textbf{(b)} sequential, \textbf{(c)} sampling. The robot is shown at the placement configuration before opening its fingers. Successful executions are outlined in green and failures in red.}
	\label{fig:execs}
	\vspace{-16pt}
\end{figure}

Fig.~(\ref{fig:benchmark_cost}) shows the placement likelihoods from Eq.~(\ref{eq:h_target}) for each of the 30 executions for all 3 methods. The likelihood encodes the closeness of the solved placement configuration to the target configuration with a likelihood of 1 when the placement configuration aligns with the target and the value approaching 0 as the distance increases. We set the placement likelihoods of unsuccessful executions to 0.
 We see from the plot that the joint method generally produces solutions closer to the target than the baselines. Most failures of the joint method also fail for the baselines. We observe the leading cause of failure being the object shifting during grasp and transfer. Fig.~(\ref{fig:execs}) shows example placement executions for each method considered in each place scene.

\subsubsection{Pick from clutter}
\begin{figure}
	\centering
	\begin{subfigure}{0.49\columnwidth}
	\includegraphics[width=\columnwidth, trim={120 0 80 150}, clip]{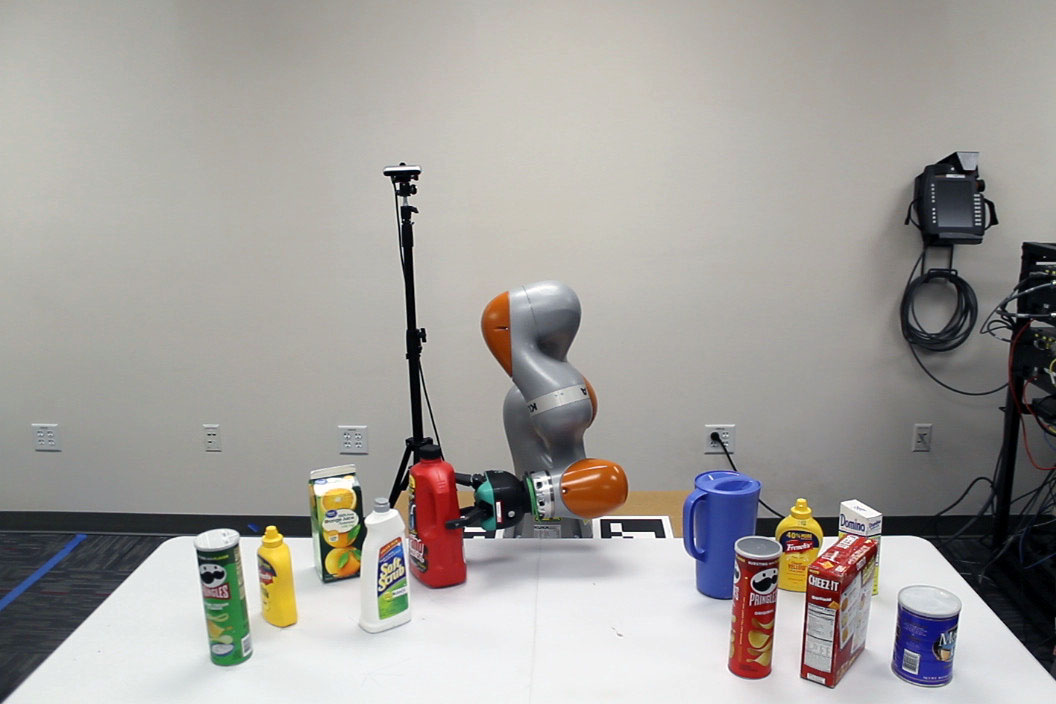}
	\label{fig:clutter_pick}
	\end{subfigure}
	\begin{subfigure}{0.49\columnwidth}
	\includegraphics[width=\columnwidth, trim={120 0 80 150}, clip]{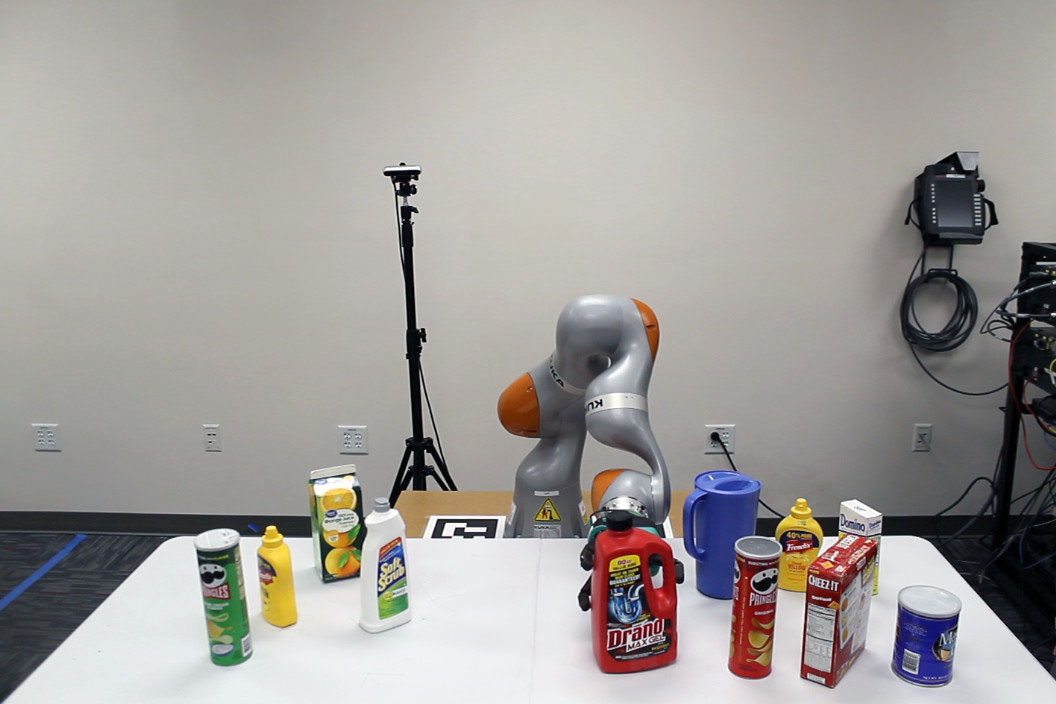}
	\label{fig:clutter_place}
	\end{subfigure}\vspace{-1em}
	\caption{Example of picking from clutter (left) and packing (right).}
	\label{fig:pack_exec}
	\vspace{-18pt}
\end{figure}

If $Z_E$ in Eq.~(\ref{eq:opt}) includes the point cloud of  nearby objects in the grasp scene, our framework can enable grasping in clutter. We experimentally validate this by having the robot pick up an object from clutter and place it as close as possible to another cluster of objects in the place scene as shown in Fig.~(\ref{fig:pack_exec}). We use the packing cost from Eq.~(\ref{eq:h_pack}) for placement. We drop the sampling-based inference for this benchmark, as the grasp sampler failed to reliably find grasps not in contact with the grasp scene clutter.

\begin{figure}
	\centering
	\includegraphics[width=\columnwidth]{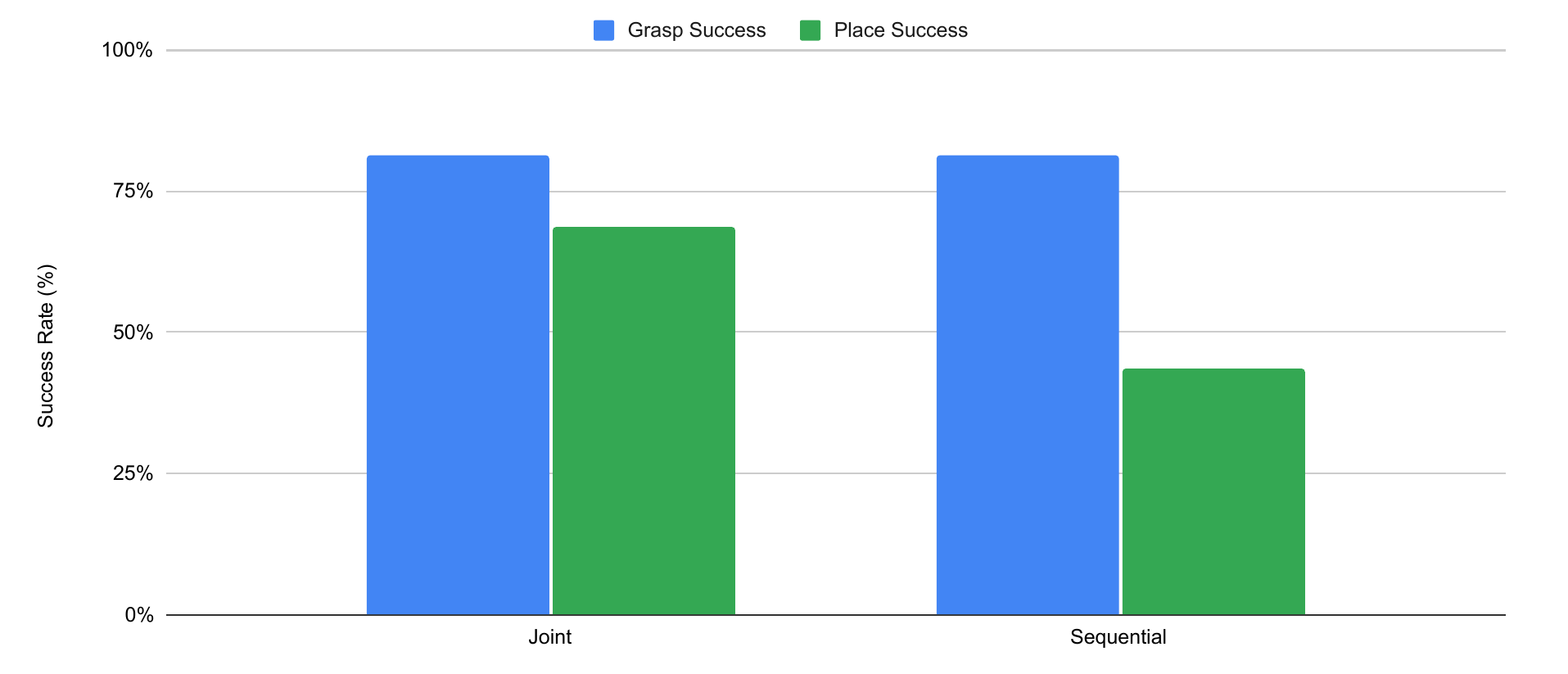} \vspace{-1.5em}
	\caption{Grasp and place success rates for pick from clutter and packing task for both joint and sequential inference methods.}
	\label{fig:pack_suc}
	\vspace{-12pt}
\end{figure}

	Similar to the previous task, we report the grasp and place success rates in Fig.~(\ref{fig:pack_suc}), 16 executions each for the joint and sequential methods. In addition to the previous requirements for grasp success, a successful grasp must not knock over any objects in the scene during grasp and lift. Fig.~(\ref{fig:pack_suc}) shows that though the grasp and place success rates drop relative to the previous experiments, joint inference still significantly outperforms the sequential baseline, with 69\% of placements being successful. The drop in success rate compared to the previous benchmark can be explained by the added difficulty of grasping from clutter for both methods.

\begin{figure}
	\vspace{3pt}
	\centering
	\includegraphics[width=\columnwidth, trim={10 0 10 20}, clip]{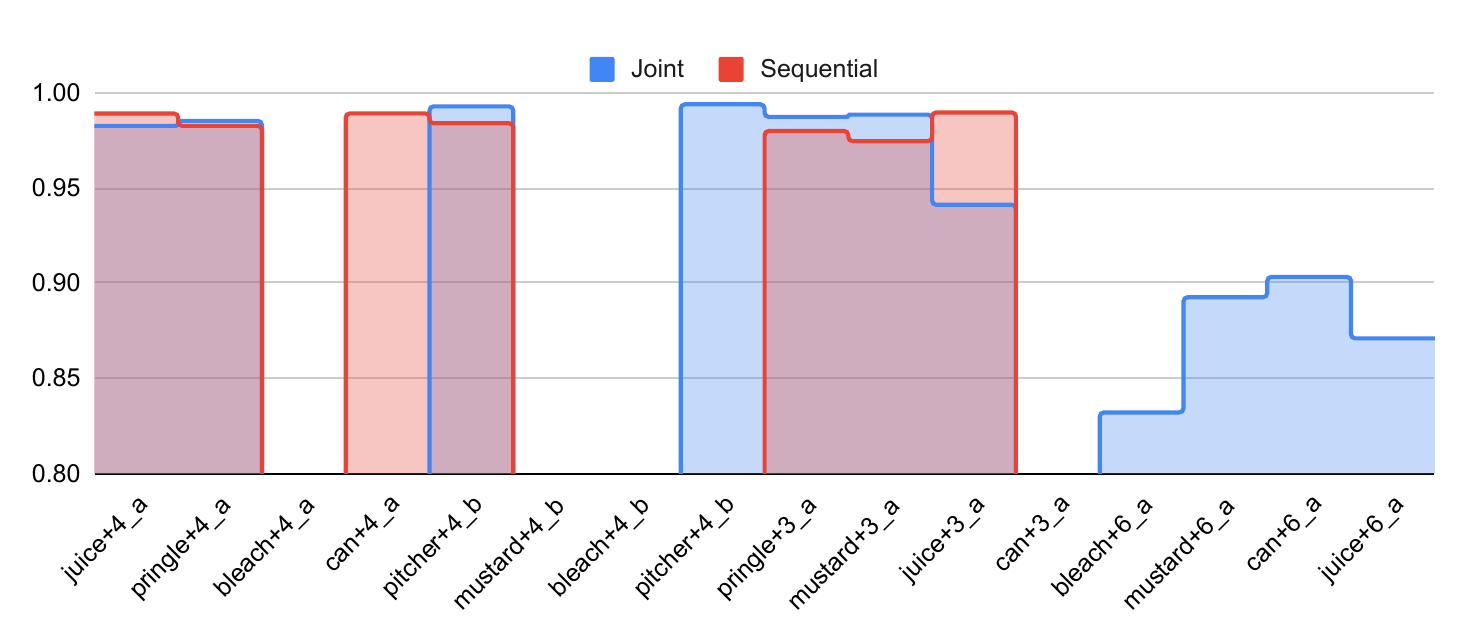}\vspace{-0.3em}
	\caption{Placement likelihood for the picking from clutter and packing task, for joint and sequential inference methods.} \label{fig:pack_cost}
	\vspace{-12pt}
\end{figure}

 We also report the packing likelihood from Eq.~(\ref{eq:h_pack}) in Fig.~(\ref{fig:pack_cost}). Here the likelihood encodes the growth in area of the bounding box enclosing all objects after placement. A likelihood of 1 denotes no growth and a likelihood of 0 denotes the bounding box has grown infinitely. We report the likelihood as 0 in cases of failed placement. Fig.~(\ref{fig:pack_cost}) shows that sequential inference performs close to joint in cases with low levels of clutter in the placement scene, but  outright fails with denser clutter. Hence the joint method is more capable of handling clutter in both grasp and place scenes.
 
 For this task the joint method had an average runtime of 82.01s (28.94 std dev). The sequential method took on average 73.59s (39.74s. std dev). We attribute the lower average to the placement optimization failing in many cases.

\subsection{Qualitative Demonstrations}

\begin{figure*}
	\vspace{0.5em}
	\centering
	\begin{subfigure}{\textwidth}
		\includegraphics[width=0.245\textwidth, trim={120 20 80 220}, clip]{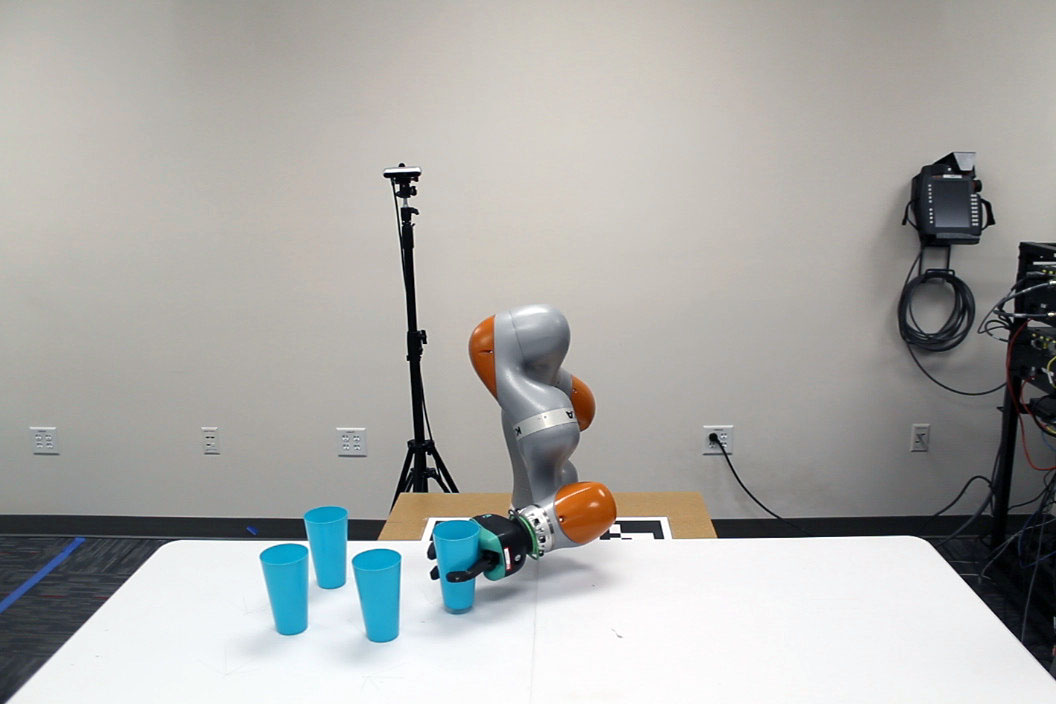}
		\includegraphics[width=0.245\textwidth, trim={120 20 80 220}, clip]{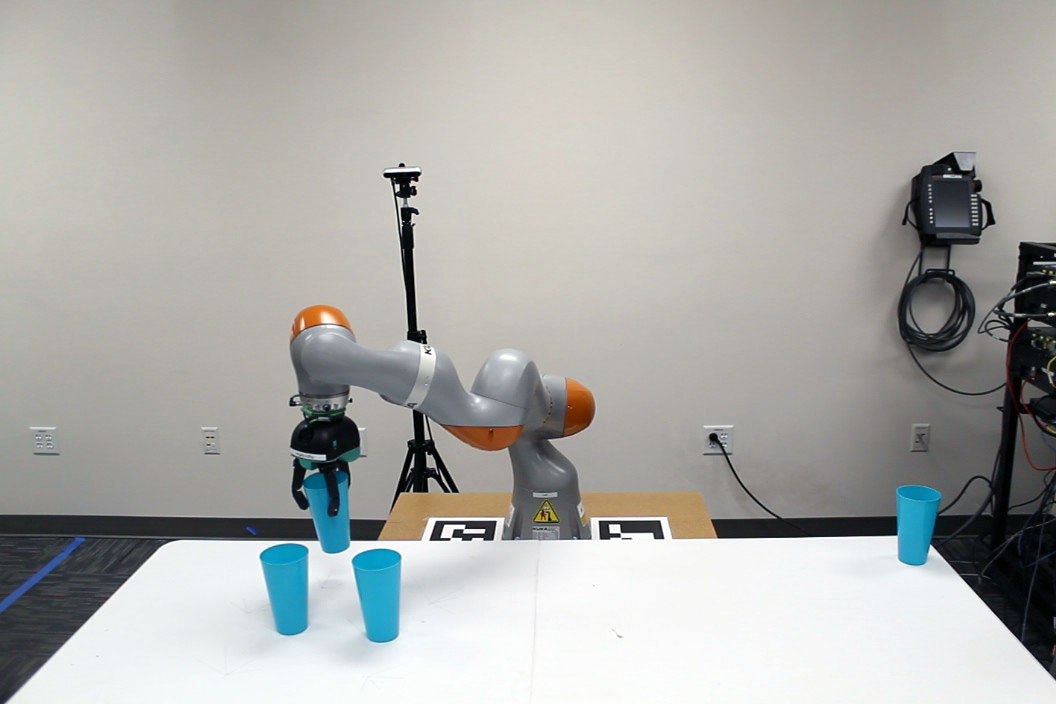}
		\includegraphics[width=0.245\textwidth, trim={120 20 80 220}, clip]{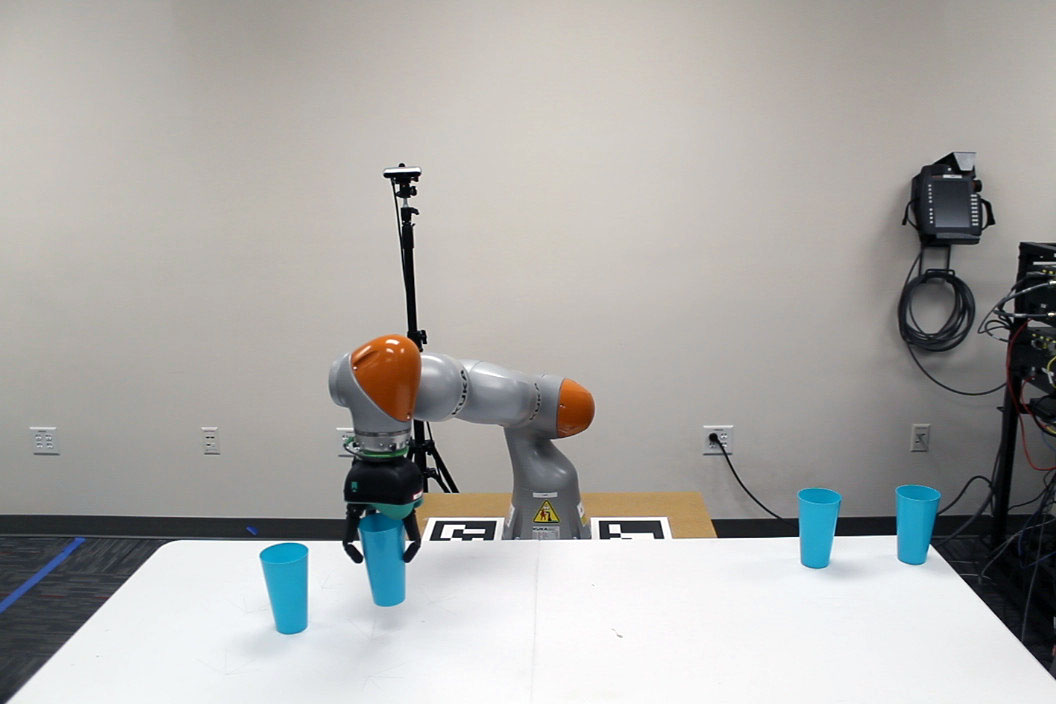}
		\includegraphics[width=0.245\textwidth, trim={120 20 80 220}, clip]{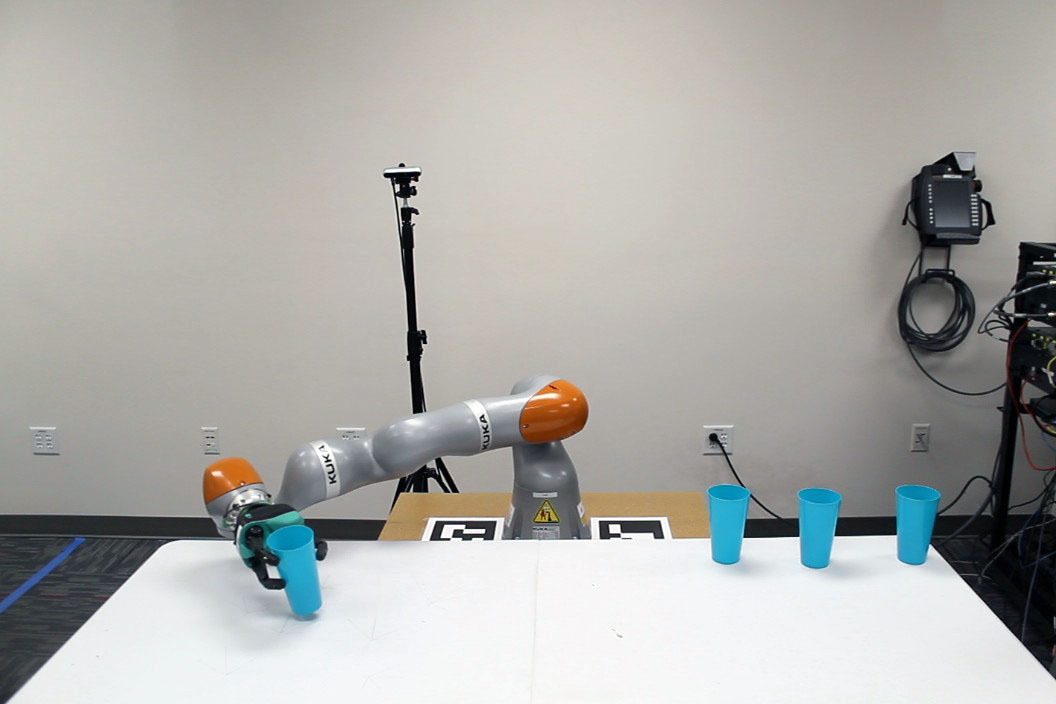}
		\label{fig:inline_grasps}
	\end{subfigure}\vspace{-1em}
	\begin{subfigure}{\textwidth}
		\includegraphics[width=0.245\textwidth, trim={120 20 80 220}, clip]{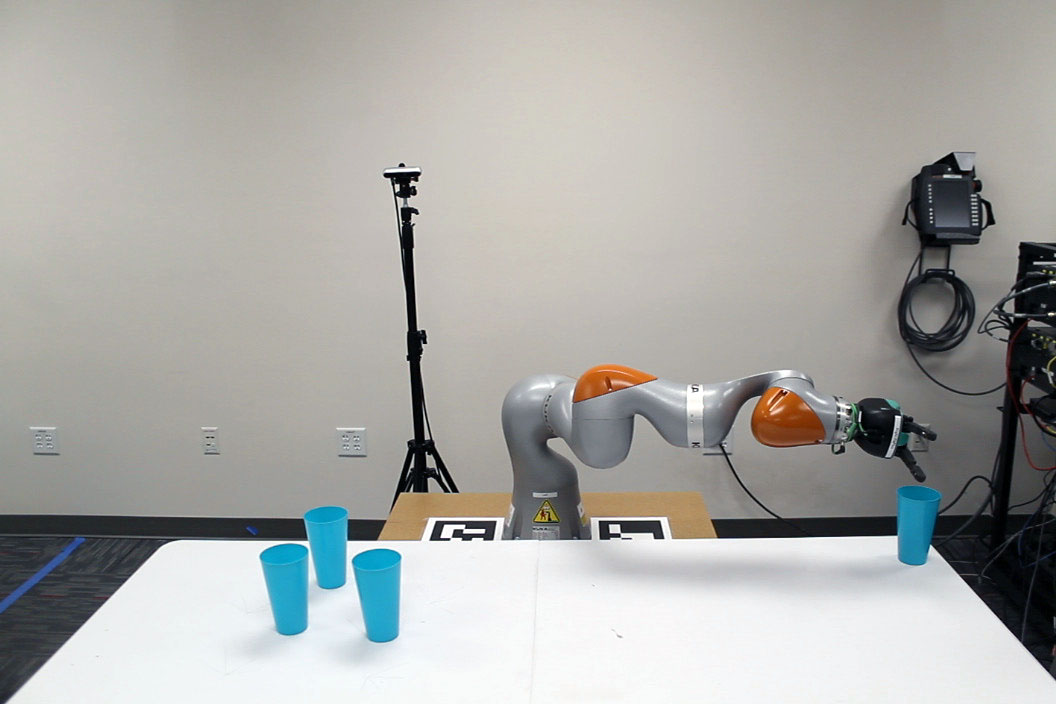}
		\includegraphics[width=0.245\textwidth, trim={120 20 80 220}, clip]{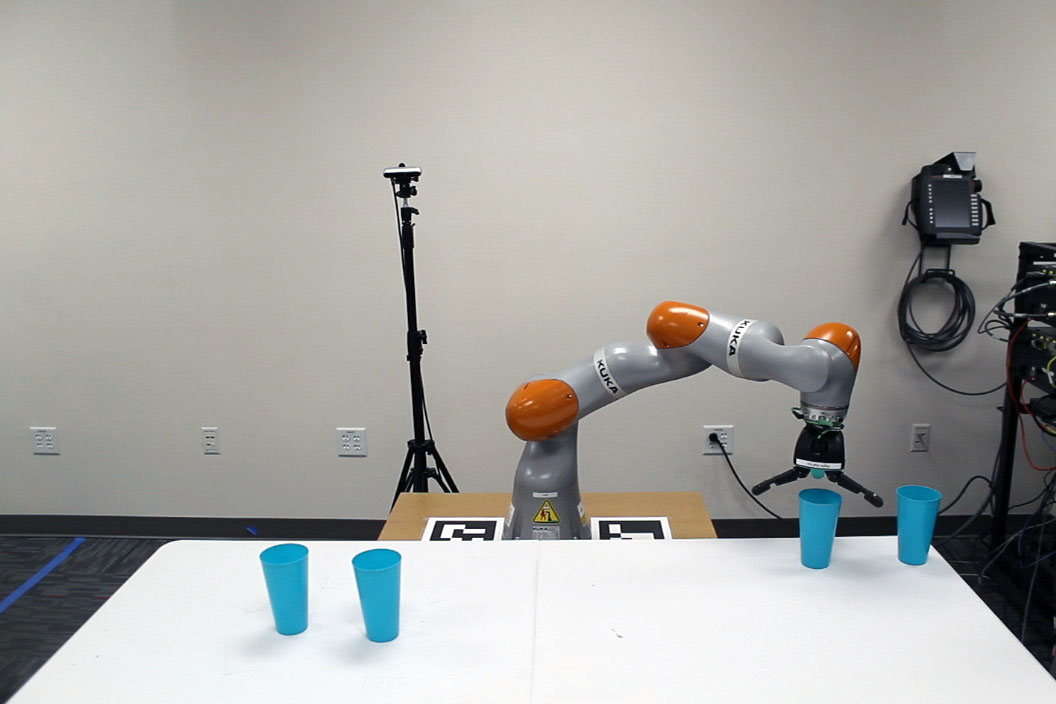}
		\includegraphics[width=0.245\textwidth, trim={120 20 80 220}, clip]{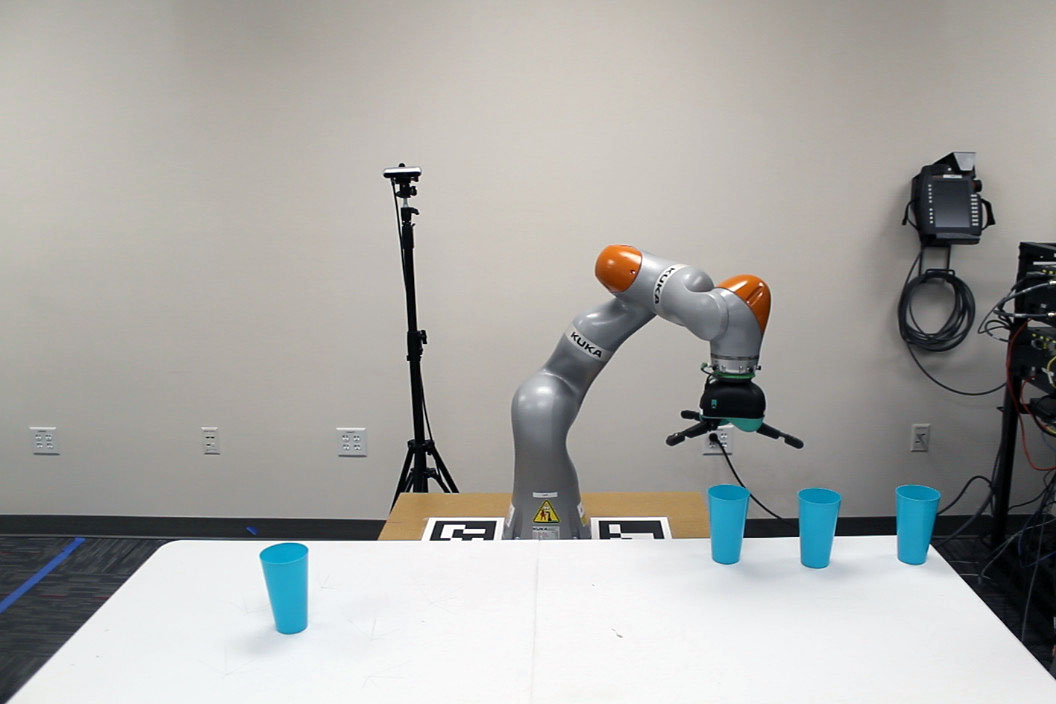}
		\includegraphics[width=0.245\textwidth, trim={120 20 80 220}, clip]{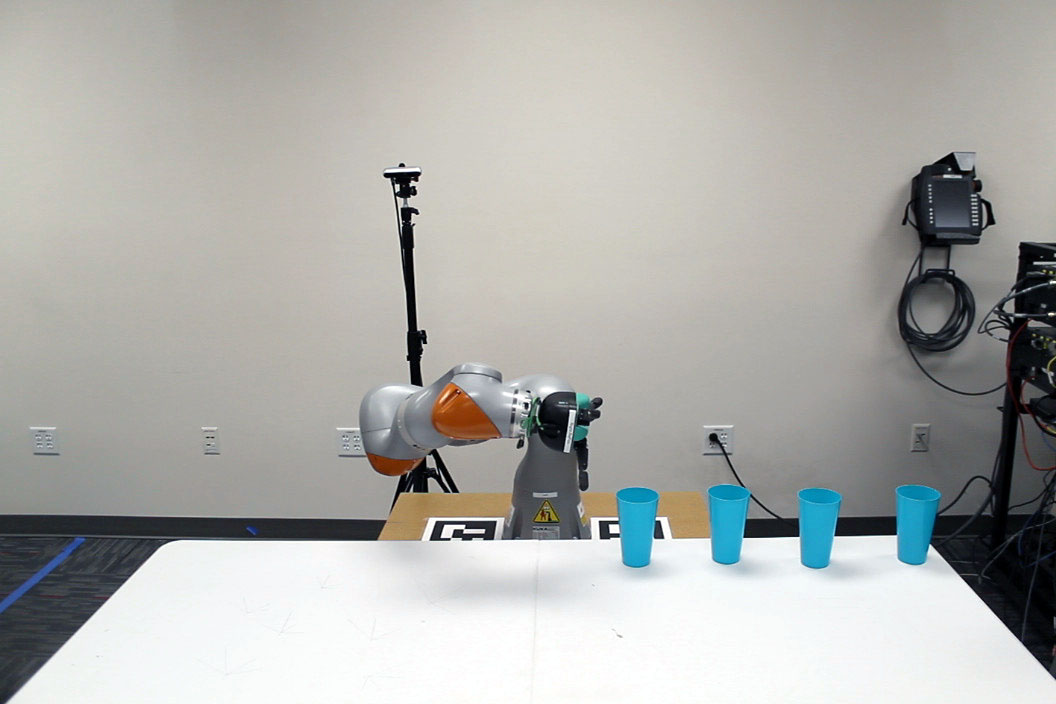}
		\label{fig:inline_placements}
	\end{subfigure}\vspace{-1em}
	\caption{Sequential pick (top row) and place (bottom row) of objects inline, using the cost defined in Eq.~(\ref{eq:h_inline}).}\label{fig:demo_inline}
\end{figure*}

\begin{figure*}
	\centering
	\begin{subfigure}{0.245\textwidth}
		\includegraphics[width=\textwidth, trim={220 30 220 220}, clip]{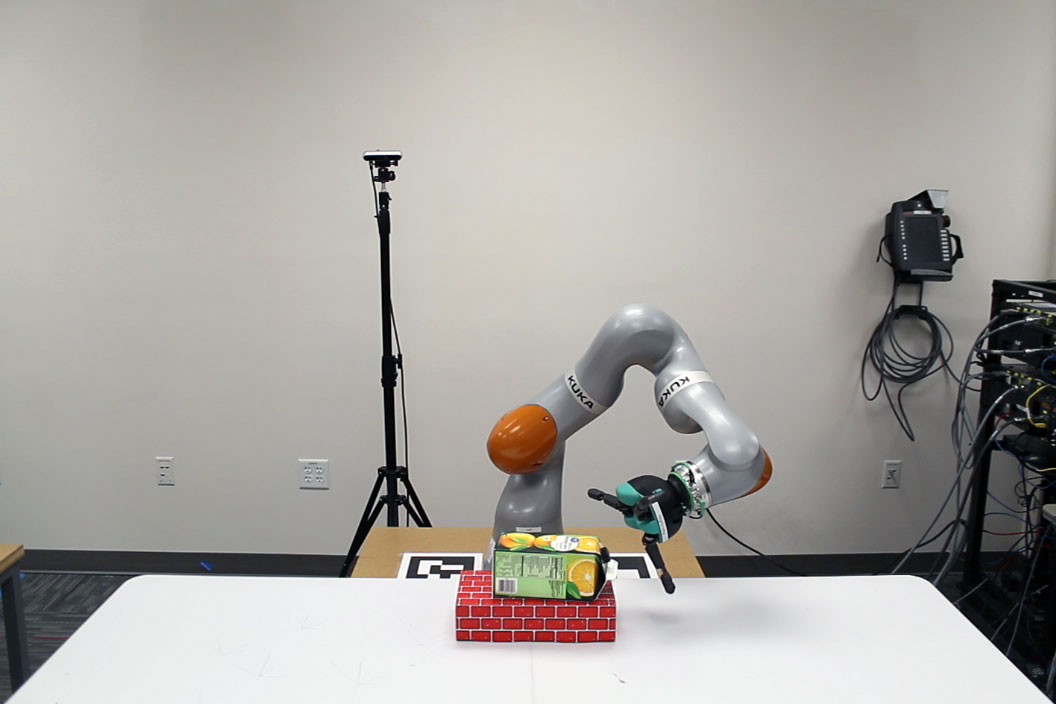}
	\end{subfigure}
	\begin{subfigure}{0.245\textwidth}
		\includegraphics[width=\textwidth, trim={220 30 220 220}, clip]{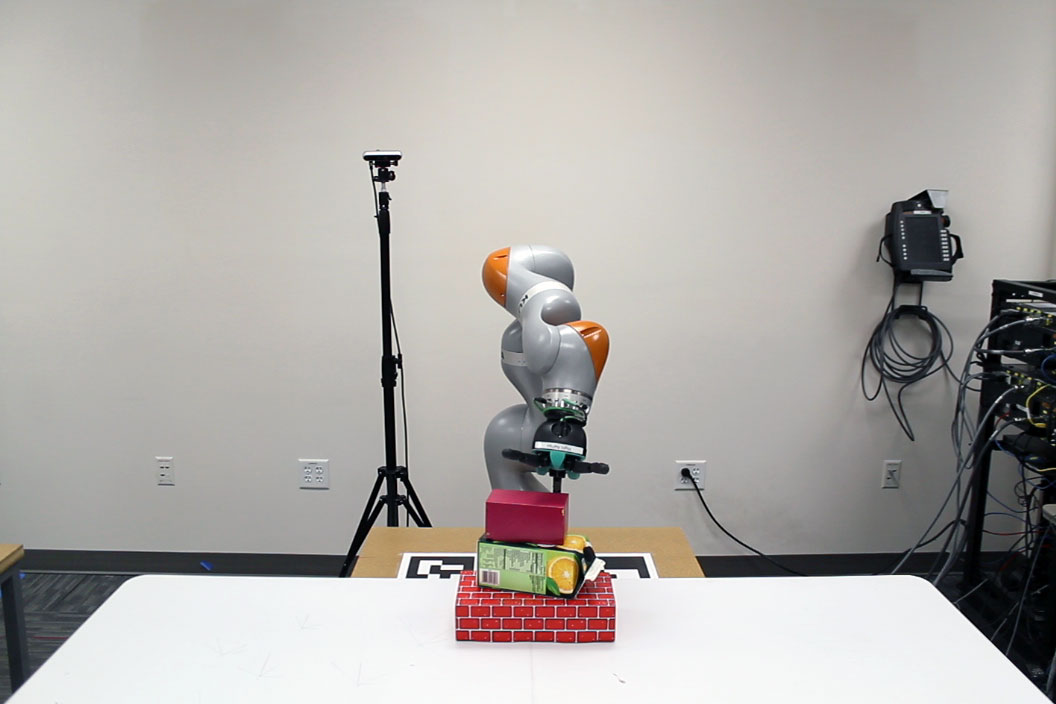}
	\end{subfigure}
	\begin{subfigure}{0.245\textwidth}
		\includegraphics[width=\textwidth, trim={220 30 220 220}, clip]{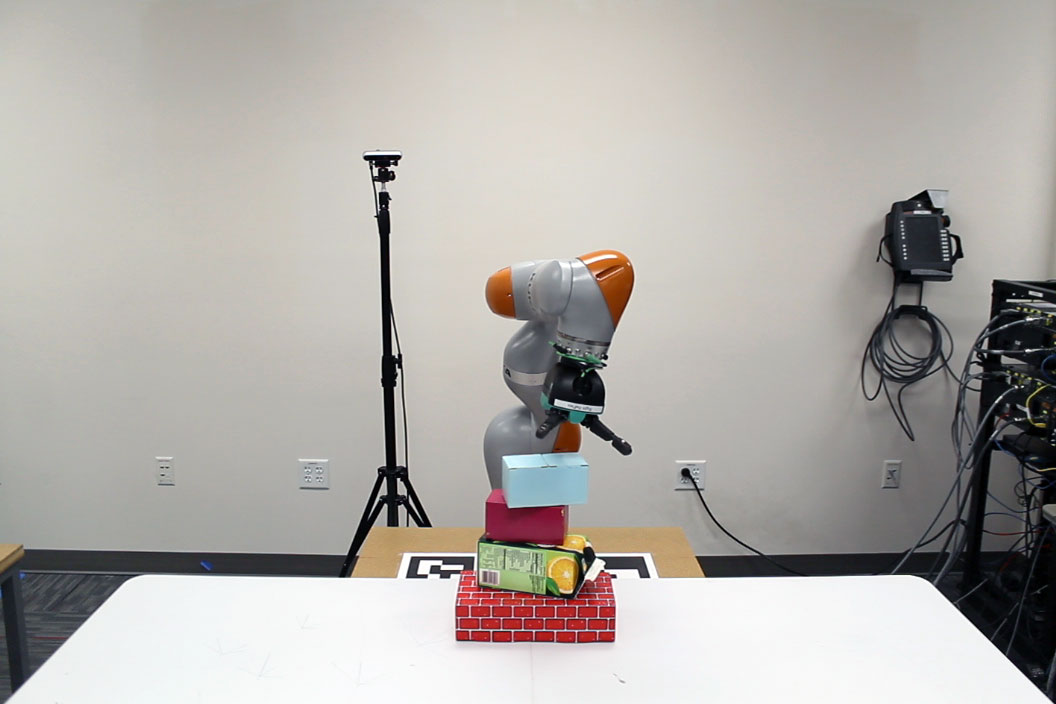}
	\end{subfigure}
	\begin{subfigure}{0.245\textwidth}
		\includegraphics[width=\textwidth, trim={220 30 220 220}, clip]{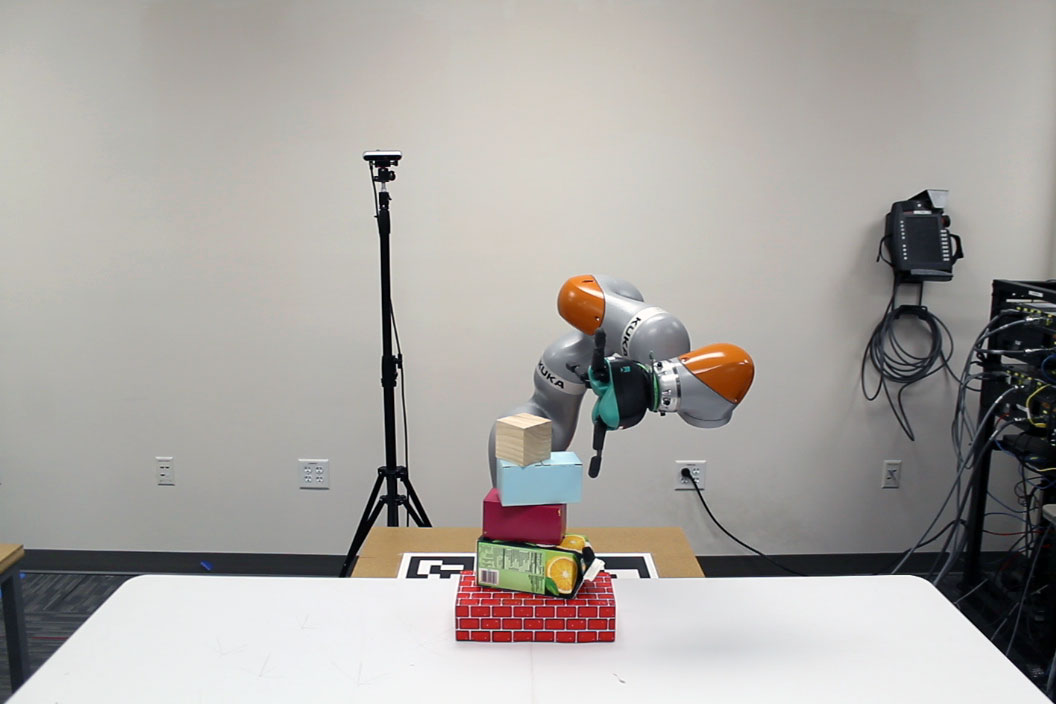}
	\end{subfigure}
	\caption{Stacking blocks in $SE(3)$ by optimizing with the placement cost in Eq.~(\ref{eq:h_stack}) which generates both top and side grasps.}\label{fig:demo_stack}
	\vspace{-10pt}
\end{figure*}

We demonstrate our joint pick and place inference in two sequential object placement tasks using costs from~\ref{sec:place_heuristics}:
\begin{enumerate}[wide, labelwidth=!, labelindent=0pt]
	\item \textbf{Place objects in line:} Using the cost from Eq.~(\ref{eq:h_inline}), we execute the solutions from our framework for the robot to rearrange a set of cups in a straight line shown in Fig.~(\ref{fig:demo_inline}).
	\item \textbf{Stacking in 6 DOF:} Fig.~(\ref{fig:demo_stack}) shows the robot stacking a sequence of blocks on top of each other, by allowing rotation in all 3-axes, each block is placed on top of another to have the minimum height possible. Thus showing our framework is capable of solving for placement configurations in $SE(3)$
\end{enumerate}

\section{Conclusion}
\label{sec:conclusion}
We presented an approach for planning a grasp for picking an unknown object jointly with a downstream placement task. By formalizing this problem as a joint inference, we were able to leverage both model-based geometric and learning-based costs and constraints into a single framework.

Many opportunities exist for future work. One could learn a post-grasp classifier, akin to the grasp classifier, in order to handle placement on non-planar surfaces or other downstream tasks (e.g. handover). Using visual or tactile feedback during placement could account for shifts of object pose relative to the gripper during transport.

In conclusion, our work is the first to show unified planning of a multi-fingered grasp for pick and place operations. Our results show the benefit of taking the placement location into account when planning grasps. In particular, we enable higher success for placement in cluttered scenes relative to planning placements sequentially after a successful grasp. We also show that our method applies to grasping in clutter scenarios without much loss in performance.
\\


\bibliographystyle{IEEEtran}
{\footnotesize
  \vspace{-1em}
 \bibliography{Placements,Packing,Grasping,General}
}
\end{document}